\documentclass{article}

\newif\ifshowrevisions
\showrevisionsfalse

\PassOptionsToPackage{numbers,sort&compress}{natbib}
\usepackage[main,final]{neurips_2026}

\usepackage[utf8]{inputenc}
\usepackage[T1]{fontenc}
\usepackage{hyperref}
\usepackage{url}
\usepackage{graphicx}
\usepackage{subcaption}
\usepackage{wrapfig}
\usepackage{placeins}
\usepackage{booktabs}
\usepackage{enumitem}
\usepackage{nicefrac}
\usepackage{microtype}
\usepackage{xcolor}
\ifshowrevisions
  \colorlet{revisionblue}{blue}
\else
  \colorlet{revisionblue}{black}
\fi
\DeclareRobustCommand{\rev}[1]{{\color{revisionblue}#1}}
\definecolor{artifactred}{RGB}{190,55,55}
\usepackage{amsmath}
\usepackage{amssymb}
\usepackage{amsfonts}
\usepackage{mathtools}
\usepackage{makecell}
\usepackage{pifont}
\usepackage{multirow}
\usepackage{amsthm}
\usepackage[most]{tcolorbox}
\usepackage[capitalize,noabbrev]{cleveref}
\usepackage{longtable}

\setlist[itemize]{itemsep=1pt, topsep=0pt}
\tcbset{
  colback=blue!5!white,
  colframe=blue!70!black,
  boxrule=0.8pt,
  arc=3pt,
  left=6pt,
  right=6pt,
  top=4pt,
  bottom=4pt,
  enhanced,
  breakable
}
\newcommand{\cmark}{\ding{51}}
\newcommand{\xmark}{\ding{55}}

\newcommand{\E}{\mathbb{E}}

\newcommand{\Ind}{\mathbb{I}}

\theoremstyle{plain}
\newtheorem{theorem}{Theorem}[section]
\newtheorem{proposition}[theorem]{Proposition}

\theoremstyle{definition}

\theoremstyle{remark}

\title{LLMs as MDP Designers: A Dependency-Aware Agentic Framework for Automated Robot Policy Generation}

\author{%
  Hong Je-Gal\\
  Department of AI \& Robotics\\
  Sejong University\\
  Seoul 05006 \\
  \texttt{jagrhong@sju.ac.kr} \\
  \And
  Jinwoo Kim\\
  Department of AI \& Robotics\\
  Sejong University\\
  Seoul 05006 \\
  \texttt{JinwooKim@sju.ac.kr} \\
  \And
  Chan-Bin Yi\\
  Department of AI\\
  Sejong University\\
  Seoul 05006\ \\
  \texttt{22012165@sju.ac.kr} \\
  \And
  Hyun-Suk Lee\thanks{Corresponding author.} \\
  Department of AI \& Robotics\\
  AI Robotics Institute (AIRI)\\
  Sejong University\\
  Seoul 05006\ \\
  \texttt{hyunsuk@sejong.ac.kr} \\
}

\begin{document}

\maketitle
\begin{abstract}
We investigate the automated generation of executable robot reinforcement learning (RL) policies from natural-language task descriptions.
Rather than using large language models (LLMs) as execution-time decision makers, we use them as designers that construct RL pipelines before deployment, including Markov decision process (MDP) formulation and simulator implementation.
We identify a phase-wise structure in this problem: MDP formulation is dependency-structured and sequential, whereas simulator implementation becomes modular once the formulation is fixed.
Based on this structure, we propose ARPG, a dependency-aware agentic framework that performs sequential MDP modeling, localized verification and correction, and simulator-structured implementation while externalizing intermediate artifacts.
ARPG targets a key failure mode of automated policy generation: executable code may still yield non-learnable policies when state, action, reward, and termination artifacts are not semantically aligned.
Experiments in Isaac Sim show that ARPG improves modeling correctness, environment executability, and policy-generation success compared with monolithic LLM baselines and off-the-shelf agentic systems.
Additional demonstrations on custom robotics, standard RL, and engineering-domain MDPs further suggest that the same modeling pipeline can be reused beyond the main benchmark suite.
\end{abstract}

\section{Introduction}

The recent emergence of physical artificial intelligence (AI) has led to an increase in the complexity and diversity of robotic tasks~\cite{zeng2023large,kawaharazuka2025vision}.
In modern industrial and service environments, robots must operate under changing task requirements, unstructured surroundings, and strict safety constraints~\cite{wang2025large,liu2025integrating,brunke2022safe}.
Consequently, designing robot control policies has become more costly and difficult, particularly for tasks that require long-horizon reasoning, contact-rich interactions, and adaptation to uncertainty~\cite{levine2018learning, peng2020learning}.

Reinforcement learning (RL) is one of the most promising solutions for acquiring robot policies that enable agents to optimize long-horizon objectives, such as stability, efficiency, and safety~\cite{kober2013reinforcement}.
However, despite significant progress in RL algorithms, generating robot RL policies remains challenging.
A primary bottleneck lies in formulation of the learning problem, rather than in policy optimization itself~\cite{dulac2021challenges}.
Formulating an RL problem for robotics requires careful specification of observations, actions, reward functions, and termination logic.
Even minor formulation errors can render learning infeasible, leading to unstable training or degenerate behaviors \cite{henderson2018deep}.
As a result, robot RL formulation is still largely performed manually by domain experts~\cite{andrychowicz2020learning}.
Whenever a task, robot, or environment changes, the entire design procedure must be repeated by human experts, resulting in long development cycles.
This reliance on manual formulation fundamentally limits the scalability and reproducibility of robot RL.

Recent advances in large language models (LLMs) have led to rapid progress across a wide range of domains, including mathematical reasoning, long-horizon planning, code synthesis, and problem solving in both scientific and human-centered tasks~\cite{wei2022chain,chen2021evaluating}.
These developments have motivated growing interest in integrating LLMs into complex decision-making systems beyond pure language tasks~\cite{yao2022react,driess2023palm}.
%
%
However, these approaches primarily emphasize flexible adaptation in open-ended and dynamically changing environments rather than the specialized, long-horizon, and safety-critical behaviors required in robot RL, especially in industrial domains.
As a result, directly employing LLMs as acting agents introduces substantial uncertainty and execution risk, making such approaches difficult to deploy in practice.
Instead, the advanced reasoning capabilities of LLMs are more effectively utilized in the RL formulation stage, where design decisions can be externalized as explicit artifacts.
RL formulation can be structured as a sequence of dependency-aware stages, such as observation specification, reward design, and termination logic. These stages produce artifacts that admit direct verification and localized correction, prior to policy training and deployment.

Motivated by this perspective, we present a dependency-aware agentic framework for \rev{automated, end-to-end generation of executable robot RL policies} from free-form natural language.
Drawing on an analysis of how human experts formulate and debug robot RL problems in practice, the roles of individual agents in the framework are guided by an explicit design rationale distilled from this human-driven process.
The framework treats LLM agents as designers and engineers who construct, ground, and iteratively refine complete RL pipelines, including sequential Markov decision process (MDP) formulation, domain-specific implementation, and debugging.
Rather than targeting code executability alone, ARPG explicitly aligns state, action, reward, and termination artifacts before implementation, since executable environments may still yield non-learnable policies when these artifacts are inconsistent.
This exposes intermediate artifacts for expert inspection, targeted modification, and localized re-execution without requiring full regeneration.
Consequently, the final output is a standalone RL policy that operates independently of language models at deployment time.

The contributions of this work are summarized as follows:

\begin{itemize}[leftmargin=2em]
\item We propose a dependency-aware agentic framework that enables \rev{automated}, end-to-end generation of robot RL policies without manual formulation.

\item We conceptualize the human-driven robot RL formulation as a structured, dependency-aware decision process, and distill explicit rationale to guide framework design with stage-wise verification and correction. This analysis clarifies the conditions under which localized correction improves the success of modeling and policy generation.

\item We design the framework to be explicitly expert-inspectable, i.e., exposing all intermediate formulation artifacts and implementation records in human-readable form, ensuring interpretability and enabling practical usability.

\item We validate the proposed framework through extensive experiments on manipulation and locomotion tasks in Isaac Sim, demonstrating substantial improvements in end-to-end policy generation success compared to unstructured baselines.
\end{itemize}

\section{Related Work}

Recent advances in robotics have increasingly leveraged AI to enable a wide range of robot behaviors, including perception-driven control~\cite{brohan2022rt,o2024open},
language-conditioned action generation~\cite{driess2023palm}, and generalist RL policies~\cite{reed2022generalist}.
One major line of research focuses on vision-language-action models~\cite{zitkovich2023rt,kim2024openvla} and large-scale generalist policies
that directly map sensory observations and language instructions to control actions.
These approaches demonstrate strong generalization across diverse tasks, particularly in open-ended and dynamically changing environments.
However, in most existing systems, AI models are primarily used as black-box executors that make decisions at execution time. These models implicitly encode task objectives, constraints, and control logic as internal representations.

In contrast, the formulation of robot RL problems, including the specification of states, actions, objectives, constraints, and termination conditions, has traditionally relied heavily on human experts~\cite{kober2013reinforcement,dulac2021challenges}.
Even in learning-based robotic systems, designing MDP structures and training pipelines typically requires manual engineering and domain knowledge~\cite{ter2025taxonomy,kaufmann2024survey}.
Recent studies in RL have explored the use of learning-based or language-based tools to assist with specific components, such as reward function design, observation selection, or policy initialization~\cite{chakraborty2023rebel, ma2023eureka, lee2026roboreward}. However, these efforts generally provide partial guidance rather than fully automated formulation.
Consequently, although execution and control have increasingly benefited from large models, formulation remains predominantly human-driven.

Outside of robotics, recent progress in LLMs has enabled a growing class of agentic frameworks that treat language models as structured problem solvers rather than execution-time agents~\cite{yao2022react, schick2023toolformer, yao2023tree}.
Across various domains, such as mathematical reasoning and optimization, these frameworks decompose complex tasks into explicit intermediate steps, iteratively refine partial solutions, and correct errors through structured feedback~\cite{wei2022chain,yao2023tree, shinn2023reflexion,madaan2023self,le2022coderl, yang2023large}.
These developments suggest that LLMs can serve as designers or engineers that construct and refine problem formulations, rather than merely producing final outputs.


Several recent studies have explored automated formulation from natural language descriptions, providing important foundations for this line of research.
Autoformalism~\cite{astorga2024autoformulation} and Optimus~\cite{ahmaditeshnizi2024optimus} both aim to translate free-form specifications into formal optimization models using LLMs.
However, these approaches are inherently limited to deterministic problems, such as linear programming and mixed-integer nonlinear programming, rather than stochastic problems.
Consequently, they cannot address the stronger dependencies induced by temporal dynamics in MDP-based formulations.



\section{Phase-wise Analysis of RL Policy Generation Problem}
\label{sec:method}
\begin{figure}[h]
    \centering
    \includegraphics[width=1.0\linewidth]{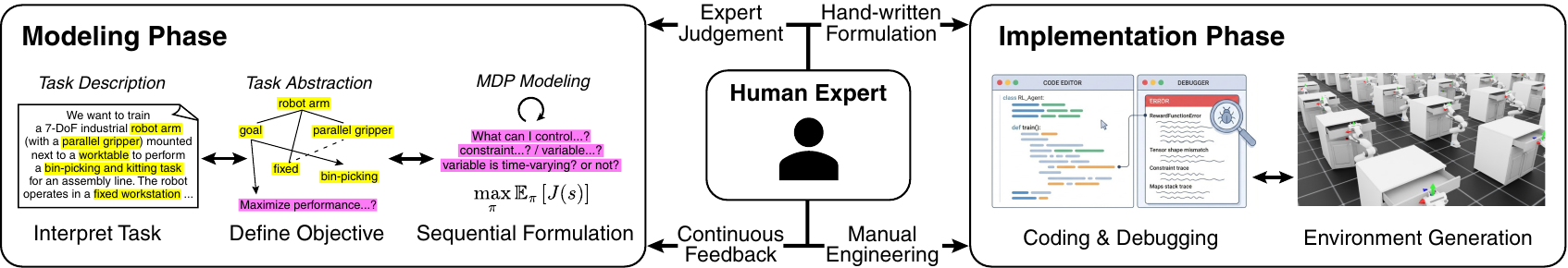}
    \caption{
    Human-driven RL policy generation process.
    The arrows illustrate iterative dependencies and back-and-forth adjustments across stages.
    }
    \label{fig:Human}
    
\end{figure}

Robot RL policy generation has traditionally relied on human experts who translate task intent and robot knowledge into a learnable control problem.
We view this process as decomposable into two coupled phases:
a \emph{modeling phase}, which interprets the task and specifies the corresponding MDP, and an \emph{implementation phase}, which realizes the MDP as an executable simulator environment and training pipeline.
Figure~\ref{fig:Human} illustrates this conceptual decomposition and the interaction between the two phases, as well as the role of human expertise in bridging them.
Although human experts often iterate across these phases through simulation feedback and debugging, the two phases have different dependency structures.

The modeling phase transforms a real-world task into a formal RL problem.
The key property of this phase is sequential dependency.
To define a valid objective or reward, the expert must first identify task parameters, decision variables, state variables, and feasibility constraints. 
Similarly, constraints cannot be specified independently of the variables and robot capabilities they restrict.
Thus, formulation artifacts are not independent modules; each artifact constrains the feasible choices of later artifacts.
We write this dependency-structured formulation process as
\begin{equation}
x \;\rightarrow\; y_1 \;\rightarrow\; y_2 \;\rightarrow\; \cdots \;\rightarrow\; y_L \;\rightarrow\; z,
\label{eq:sequential_modeling}
\end{equation}
where $x$ is the task description, $\{y_i\}_{i=1}^L$ are intermediate formulation artifacts such as parameters, variables, objectives, and constraints, and $z$ is the complete MDP formulation.
This structure makes early formulation errors especially costly, because they can propagate to later artifacts and eventually yield an invalid or non-learnable MDP.

Unlike the modeling phase, the implementation phase has a modular structure rather than a semantic sequential dependency structure.
Once $z$ is fixed, this phase translates the MDP into an executable learning pipeline by instantiating it in a target simulator.
In manager-based environments such as Isaac Lab, this instantiation is modular: scene, action, observation, event, termination, reward, and policy configurations are composed into an executable environment.
Although some modules have explicit simulator-level dependencies, many can be generated independently after the MDP formulation is fixed.
We therefore view implementation as a modular transformation
\begin{equation}
z \;\rightarrow\; g(z) = \{\, g_1(z),\; g_2(z),\; \ldots,\; g_M(z) \,\},
\label{eq:parallel_implementation}
\end{equation}
where each $g_m(\cdot)$ produces a simulator-specific configuration module.
In practice, these modules can be generated in simulator-dependent stages when explicit inter-module dependencies exist.

Based on this phase-wise structure, the objective of automated RL policy generation is to produce a formulation and implementation whose trained policy satisfies a task-specific success criterion:
\begin{equation}
\mathbb{E}\!\left[
\mathbb{I}\!\left( J(g(z)) \ge \tau \right)
\;\middle|\;
z = f(x)
\right],
\label{eq:final_objective}
\end{equation}
where $J(g(z))$ is the performance of the trained policy produced from the implementation $g(z)$, $\tau$ is a task-specific requirement, $f$ denotes the sequential formulation process in Equation~\eqref{eq:sequential_modeling}, and $g$ denotes the simulator implementation process in Equation~\eqref{eq:parallel_implementation}.
This objective captures the design goal of generating an MDP formulation and implementation that lead to a task-valid learned policy, while reflecting the different dependency structures of modeling and implementation.

\section{ARPG: Agentic LLM-Based Framework for Automated Robotics Policy
Generation}

\subsection{Framework Overview and Design Rationale}
\label{subsec:framework_structure}

We propose ARPG, an agentic framework for automated robot policy generation using multi-agent LLMs.
ARPG is designed to reflect the phase-wise structure identified in Section~\ref{sec:method}: dependency-aware sequential modeling followed by simulator-structured modular implementation.
For MDP modeling, parallel generation can break dependency chains among parameters, variables, objectives, constraints, and rewards \cite{kim2025towards}, while monolithic generation makes intermediate decisions opaque and difficult to correct~\cite{leerefinebench}.
A further challenge in iterative refinement is that fixing one component may inadvertently invalidate others.
We refer to this phenomenon as \emph{error reintroduction}.
Unbounded error reintroduction leads to unstable refinement, motivating localized correction with explicit artifact boundaries.

\begin{figure}[h]
    \begin{center}
    \centerline{\includegraphics[width=0.95\linewidth]{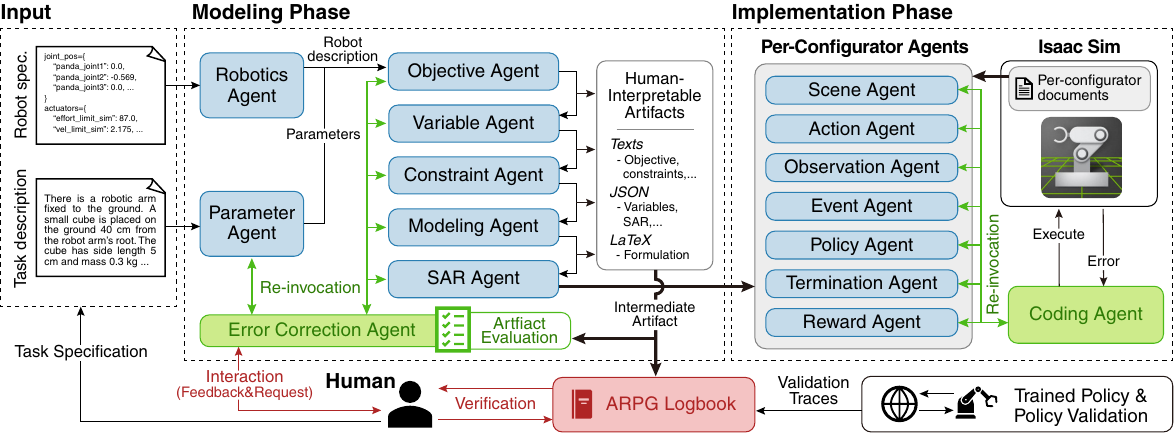}}
    \caption{
    Overview of the ARPG framework for structured formulation and policy generation.
    }
    \label{fig:ARPG_flow}
    \end{center}
    \vspace{-1.5em}
\end{figure}

To address the formulation-side challenges, we introduce the following design principles:
1) \emph{dependency-respecting sequential formulation},
2) \emph{explicit externalization of formulation artifacts}, and
3) \emph{stage-wise verification with localized correction to \rev{limit} error reintroduction}.
In addition, for implementation, we consider simulator-structured modular generation, exploiting parallelism where configuration modules have no unresolved dependencies.
Together, these choices align the agentic structure with the sequential nature of MDP formulation and the modular structure of simulator implementation.

Figure~\ref{fig:ARPG_flow} illustrates the overall structure of the proposed framework.
Given a natural-language task description and a robot specification, the framework first constructs a formal MDP through a modeling phase. Then, it generates an executable Isaac Lab training pipeline through an implementation phase.
All intermediate artifacts produced during this process, such as texts, JSON records, and LaTeX scripts, are explicitly materialized and stored in an ARPG logbook. This artifact log supports expert inspection, partial re-execution, and targeted modification without regenerating the entire pipeline, although such human intervention is not used in our automated evaluation.

\subsection{Modeling Phase}
The modeling phase requires two types of input.
The first is a free-form task description provided by a human user,
which specifies the desired behavior, success criteria, and operating conditions.
\rev{The second is a Python or USD file describing a robot already integrated into the target simulator.}
This file encodes the robot's kinematic structure, actuators, joints,
and physical constraints.
Together, these inputs define the task intent and the feasible control space of the agent.

These two inputs are processed by different specialized agents.
The task description is processed by a parameter agent that extracts task-specific parameters, such as object properties and workspace settings. This agent provides the semantic basis for identifying decision variables, objectives, and constraints.
The robot specification is processed by a dedicated robotics agent. This agent parses the code file and translates simulator-level representations into a concise natural-language summary.
This summary reveals the controllable components, admissible motions, and relevant physical limitations of the robot. This allows subsequent modeling agents to reason about the robot's capabilities without interacting directly with simulator-specific formats. It also enables the modeling phase in the framework to operate across different robot types, with only a change in the robot description.

After this input processing, the core formulation of the MDP proceeds through a fixed and strictly ordered sequence of specialized agents, in accordance with the rationale of dependency-respecting sequential formulation.
Based on the extracted parameters and the robot capability description, first, an objective agent specifies the optimization objective.
Then, a variable agent identifies the decision variables and state variables, which are required to characterize the objective.
A constraint agent defines the feasibility conditions that restrict valid actions and state transitions based on the identified variables.
The objective, variables, and constraint specifications are then formalized into mathematical expressions by a modeling agent.
Finally, a state-action-reward (SAR) agent integrates these elements to produce a complete MDP formulation $z$.

We illustrate this sequence with a simplified Humanoid locomotion example below. More detailed task descriptions and artifact traces for all tasks are provided in Appendix \ref{app:representative_logbook_trace} and the supplementary material, respectively.
Notably, all the intermediate artifacts during the modeling phase are provided in a textual form (e.g., objective, constraints), JSON records (e.g., variables, SAR structure), or LaTeX scripts (e.g., MDP formulation), making intermediate decisions inspectable and locally revisable.
\begin{tcolorbox}[
  title=Running example: Humanoid locomotion,
  colback=gray!5!white,
  colframe=gray!60!black,
  boxrule=0.6pt,
  arc=2pt,
  left=4pt,
  right=4pt,
  top=3pt,
  bottom=3pt
]
\footnotesize
For ``make the humanoid walk forward without falling,'' ARPG constructs the MDP through an artifact chain:
\begin{itemize}[leftmargin=1.2em, nosep]
  \item \textbf{Parameters}: control interval, 16s horizon, $+X$ forward direction, base-height fall threshold, initialization ranges, and actuator limits.
  \item \textbf{Robot summary}: actuated joints, actuator semantics, feasible commands, base link, contact links, and physical limits.
  \item \textbf{Objective}: forward locomotion with upright posture, heading alignment, lateral stability, and fall avoidance.
  \item \textbf{Variables}: joint states, joint command targets, base pose and velocities, base height, heading error, uprightness, forward/lateral motion, and contact signals.
  \item \textbf{Constraints}: timing, action decimation, actuator bounds, orientation consistency, contact consistency, timeout termination, and fall termination.
  \item \textbf{SAR}: proprioceptive state, continuous joint-level action space, and reward combining forward progress, stability, lateral-drift, and fall-penalty terms.
\end{itemize}
\end{tcolorbox}

The sequential execution of the modeling agents can be formalized according to the process in Equation~\eqref{eq:sequential_modeling} as
\begin{equation}
\begin{aligned}
z &= f(x)
   = f_L \circ f_{L-1} \circ \cdots \circ f_1 (x),
\end{aligned}
\end{equation}
where $f_i$ denotes the $i$-th stage-specific formulation transformation that produces the intermediate formulation artifact $y_i = f_i\!\left(x,\{y_j\}_{j=1}^{i-1}\right)$.
This explicit externalization of formulation artifacts
ensures that intermediate decisions are materialized,
inspectable, and independently revisable.
This separation explicitly distinguishes formulation artifacts
from the transformations that generate them,
eliminating ambiguity in the sequential modeling process.


To maximize the effectiveness of this externalization, the framework introduces an error correction (EC) agent to limit error reintroduction in the sequential process.
{\color{revisionblue}
Each agent in the modeling phase produces both a formulation artifact and an associated self-assessed confidence score in $[0,1]$.
When this score falls below a predefined threshold, the EC agent evaluates the artifact using agent-specific checklists and returns a localized diagnosis.
For example, the parameter agent's checklist examines task-critical values and bounds, consistency of units and scales, and potential contradictions.
The responsible modeling agent is then re-invoked with its previous artifact, the diagnosis, and the same upstream inputs.
This correction cycle is repeated within a configured retry limit while accepted upstream artifacts remain fixed, enabling localized revision of the artifacts without re-executing the entire modeling process.
Further analysis of EC invocation patterns and correction outcomes is provided in Appendix~\ref{app:modeling_correction_dynamics}.
}

\subsection{Implementation Phase}

After the modeling phase produces an MDP formulation $z$, ARPG instantiates it as an executable training pipeline in the target simulator.
The implementation structure depends on the simulator backend.
For Isaac Lab, ARPG exploits its manager-based design, where scene, policy, action, observation, event, termination, and reward configurations are composed into a single environment.
This modular structure allows ARPG to generate independent components in parallel while respecting explicit inter-module dependencies, such as shared scene definitions or observation specifications.

ARPG represents this process as simulator-structured staged generation:
\begin{equation}
g(z;\mathcal{D})
=
G_N\!\circ G_{N-1}\!\circ \cdots \circ G_1(z;\mathcal{D}),
\qquad
G_k=\{g^{(k)}_i\}_{i\in\mathcal{I}_k}.
\label{eq:staged_implementation}
\end{equation}
Here, $G_k$ denotes the set of configuration-module generators executed at stage $k$.
Generators within the same $G_k$ can run in parallel, while the composition across stages captures explicit simulator-level dependencies.
For Isaac Lab, we use three stages ($N=3$): global scene and policy specifications; scene-dependent action, observation, event, and termination configurations; and reward configurations that depend on scene and observation specifications.

After module generation, a coding agent aggregates the resulting artifacts into executable environment and training code.
If compilation or runtime errors occur, the error trace is analyzed to localize the failure to a specific configuration stage or module, and ARPG selectively re-invokes the responsible configuration agent for localized correction.
When errors arise from cross-module interactions or inconsistent assumptions, the coding agent performs a coordinated revision of the assembled code.
Upon successful execution, the pipeline yields both an executable simulation environment and a trained RL policy.

\subsection{Common Artifact Management and Logbook}
Beyond the phase-specific agents, the ARPG framework provides a shared artifact management mechanism that governs how intermediate results are externalized, modified, and reused across stages. All intermediate artifacts, including formulation artifacts, configuration specifications, generated code fragments, and policy validation traces, are systematically collected in the ARPG logbook.
This representation explicitly records stage-wise dependencies and configuration boundaries. This enables both localized modification of individual components and global revision of the assembled code.
Human users can directly inspect and modify specific artifacts when the framework is used as an expert-facing design tool, while the same artifact boundaries are also used by agents to support targeted re-invocation in the automated pipeline.
This supports efficient iterative refinement with minimal recomputation, independent of the underlying simulator-specific stage structure.

\subsection{Theoretical Justification}
\label{subsec:theoretical_justification}

We provide a \rev{conditional explanatory model} for the proposed ARPG framework, which consists of the modeling phase and the implementation phase.
This \rev{model} focuses on the tailored sequential agentic structure of the
modeling phase, which determines the correctness of the resulting formulation.
In the framework, end-to-end success occurs if and only if both phases succeed
\begin{equation}
\mathbb{P}(S_{\mathrm{ARPG}})
= \mathbb{P}(S_M)\,\mathbb{P}(S_I \mid S_M),
\end{equation}
where $S_M$ and $S_I$ denote the success events of the modeling and implementation
phases, respectively, since the implementation phase operates conditional on a completed formulation.

We consider the modeling phase that produces a set of $L$ formulation artifacts.
Each artifact is either valid or invalid at termination.
Let $A_i^{(t)}$ denote the event that formulation artifact $i$ is valid after the $t$-th verification-correction iteration.
To reflect the sequential dependency among artifacts, we interpret $p_i^{(t)}$ as the conditional residual error probability that artifact $i$ remains invalid after iteration $t$, conditioned on the preceding artifacts being valid:
\[
p_i^{(t)} := \mathbb{P}\!\left((A_i^{(t)})^c \mid A_1^{(t)},\ldots,A_{i-1}^{(t)}\right).
\]
The quantity $p_i^{(\infty)}$ denotes the corresponding conditional residual error probability at termination.
We define the modeling success event as
$S_M := \{\text{all formulation artifacts are valid at termination}\}$,
with corresponding success probability
\begin{equation}
\label{eq:sm_def}
\mathbb{P}(S_M)
=
\prod_{i=1}^{L} \bigl(1 - p_i^{(\infty)}\bigr).
\end{equation}
The ARPG framework enforces explicit externalization of formulation artifacts,
stage-wise verification, and localized correction.
These structural constraints influence the effectiveness of error correction and
the degree of unintended interference across formulation artifacts.
This intuition can be formalized as the following comparison result.

\begin{proposition}
\label{prop:relative_improvement}
Consider a baseline formulation strategy and the ARPG framework, both of which
admit artifact-wise formulation errors.
Under the \rev{assumption} that ARPG induces strictly better effective correction behavior (e.g., higher correctability or lower interference) for at least one formulation artifact, while all other artifacts are
unchanged, the modeling success probability satisfies
\[
\mathbb{P}(S_M^{\mathrm{ARPG}})
>
\mathbb{P}(S_M^{\mathrm{base}}).
\]
\end{proposition}

This proposition does not assume that alternative strategies lack error correction capabilities.
Single-agent reasoning may allow for implicit self-correction, and parallel formulation strategies may use post hoc reconciliation.
However, in practice, single-pass monolithic reasoning \rev{may exhibit weaker} effective error correction due to the lack of explicit, verifiable intermediate artifacts. Meanwhile, the parallel modification of multiple formulation artifacts \rev{may increase} cross-artifact interference.
\rev{An LLM may miss an error, fail to repair it, or introduce a new inconsistency, so the assumed correction behavior need not hold in every invocation.}
\rev{Under the proposition's assumptions,} ARPG induces a higher modeling success probability by promoting effective correction and limiting unintended interactions.
More detailed analyses are provided in Appendix~\ref{app:correction_dynamics}.

\section{Experimental Results}

\subsection{Experimental Setup}

For evaluation, we choose five main benchmark tasks that cover a variety of state dimensionality, control complexity, and interaction structure, ranging from low-dimensional control to high-degree-of-freedom locomotion and contact-rich manipulation.
All experiments are implemented in Isaac Lab (v2.1.0) on top of NVIDIA Isaac Sim (v4.5.0).

\begin{itemize}[leftmargin=2em]
\item \textit{CartPole} is a low-dimensional classic control task for evaluating basic MDP formulation with simple state and action spaces.
\item \textit{Ant and Humanoid} are high-dimensional locomotion tasks requiring coordinated long-horizon control across many degrees of freedom.

\item \textit{Reach and Lift} are manipulation tasks in which the robot must move its end-effector to a target pose and grasp and lift an object.
\end{itemize}

We evaluate ARPG to empirically assess whether dependency-aware sequential MDP formulation improves formulation correctness and translates into higher end-to-end policy generation success.
To this end, we introduce three complementary metrics that reflect different stages of the policy generation pipeline. Their detailed criteria are provided in the Appendix~\ref{app:Details_of_Success}.
\begin{itemize}[leftmargin=2em]
\item \textit{Modeling success rate (MSR)}:
The proportion of trials whose MDP components are logically correct and task-aligned for implementation\rev{, or whose trained policies
satisfy the PGSR criteria}.

\item \textit{Environment implementation success rate (EISR)}:
The proportion of trials in which the generated code executes without syntax or runtime errors in a standard Isaac Sim environment.

\item \textit{Policy generation success rate (PGSR)}:
The proportion of trials in which RL training converges and the trained policy satisfies the task-specific success criteria.
\end{itemize}

\rev{We use the following language model configurations for ARPG and the baselines.}
In ARPG, all agents use GPT-5.2.
The monolithic baselines use GPT-5.2 and Gemini-3.
We also evaluate off-the-shelf agentic systems, Claude Code with Claude Sonnet 4.6 and Codex with GPT-5.2.
\rev{All systems receive the same task description, robot/USD information, and per-configuration documentation.
This documentation is embedded directly in the prompt for monolithic baselines and provided as a text file to the off-the-shelf agentic systems.}
\rev{For all configurations that allow code revision using execution feedback, we fix the maximum number of revision rounds at five.}
Unless otherwise noted, results are evaluated over 10 independent runs.

\begin{wraptable}{r}{0.5\textwidth}
\centering
\vspace{-0.5em}
\caption{Comparison of structural components.}
\label{tab:structural_components}
\begin{small}
\setlength{\tabcolsep}{2pt}
\begin{tabular}{lccc}
\toprule
Component
& \makecell{Monolithic\\LLM}
& \makecell{Off-the-shelf\\agents}
& \makecell{ARPG\\(Ours)} \\
\midrule
\makecell[l]{Component-wise\\implementation}
& \xmark
& \cmark
& \cmark \\
\midrule
\makecell[l]{Dependency-aware\\MDP artifactization}
& \xmark
& \xmark
& \cmark
\\
\midrule
\makecell[l]{Artifact-localized\\correction}
& \xmark
& \xmark
& \cmark \\
\bottomrule
\end{tabular}
\end{small}
\end{wraptable}

Table~\ref{tab:structural_components} summarizes the architectural ingredients present in each evaluated system.
Monolithic LLM baselines may reason over the full task in a single context, but they do not externalize MDP formulation as a dependency-aware artifact chain.
Off-the-shelf agentic systems provide stronger implementation decomposition and execution-based debugging, but they do not explicitly structure MDP formulation as a dependency-ordered artifact chain.
Consequently, such systems may improve environment executability without ensuring that MDP-related artifacts are semantically aligned for policy learning.
ARPG differs by coupling component-wise implementation with dependency-aware MDP artifactization and artifact-localized correction.
This structural distinction motivates our stage-wise evaluation using MSR, EISR, and PGSR.




\subsection{Overall Performance}
\label{sec:overall}
We first evaluate ARPG against monolithic LLM baselines and off-the-shelf agentic systems across the benchmark tasks, instantiating three practical design regimes: monolithic generation, general-purpose agentic code generation, and dependency-aware MDP-centered generation.
%
%
%
\rev{The monolithic baselines test whether a single model can generate the full pipeline, with one-shot generation limited to a single pass and iterative generation allowing execution-based debugging.}
The agentic baselines test whether component-wise code generation and execution-based debugging are sufficient without dependency-aware MDP artifactization.
We report MSR, EISR, and PGSR to decompose the pipeline into formulation correctness, implementation executability, and final policy success.
This decomposition captures the central failure-mode separation targeted by ARPG: executable environments do not necessarily encode learnable, task-aligned MDPs, while valid MDPs can still fail due to simulator-specific implementation errors.

\begin{table}[h]
\setlength{\tabcolsep}{3pt}
\centering
\caption{Success rates across benchmark tasks. All results are reported as triplets in the order of MSR, EISR, and PGSR. Bold indicates the best value for each metric within a task. Corresponding 95\% Wilson confidence intervals are provided in Appendix~\ref{app:confidence_intervals}.}
\label{tab:arpg_comparison}
\begin{small}
\resizebox{\textwidth}{!}{%
\begin{tabular}{lcccccc}
\toprule
\multirow{2}{*}{Task}
& \multicolumn{4}{c}{GPT-5.2}
& Gemini-3
& Sonnet 4.6 \\
\cmidrule(lr){2-5}
\cmidrule(lr){6-6}
\cmidrule(lr){7-7}
& ARPG & One-shot & Iterative
& Codex & One-shot & Claude \\
\midrule
CartPole
& \textbf{1.0} / \textbf{1.0} / \textbf{1.0}
& 0.4 / 0.0 / 0.0
& \textbf{1.0} / \textbf{1.0} / \textbf{1.0}
& \textbf{1.0} / 0.6 / 0.6
& 0.6 / 0.0 / 0.0
& \textbf{1.0} / \textbf{1.0} / \textbf{1.0} \\
Ant
& \textbf{1.0} / \textbf{0.9} / \textbf{0.9}
& 0.2 / 0.0 / 0.0
& 0.2 / 0.3 / 0.1
& 0.4 / 0.8 / 0.4
& 0.1 / 0.0 / 0.0
& 0.0 / \textbf{0.9} / 0.0 \\
Humanoid
& \textbf{0.9} / 0.7 / \textbf{0.4}
& 0.1 / 0.0 / 0.0
& 0.1 / 0.3 / 0.0
& 0.1 / 0.5 / 0.1
& 0.3 / 0.0 / 0.0
& 0.0 / \textbf{0.9} / 0.0 \\
Reach
& \textbf{0.9} / 0.9 / \textbf{0.9}
& 0.0 / 0.0 / 0.0
& 0.0 / 0.6 / 0.0
& 0.4 / 0.3 / 0.2
& 0.0 / 0.0 / 0.0
& 0.5 / \textbf{1.0} / 0.3 \\
Lift
& \textbf{1.0} / \textbf{0.9} / \textbf{0.4}
& 0.0 / 0.0 / 0.0
& 0.0 / \textbf{0.9} / 0.0
& 0.0 / 0.6 / 0.0
& 0.0 / 0.0 / 0.0
& 0.0 / \textbf{0.9} / 0.0 \\
\bottomrule
\end{tabular}%
}
\end{small}
\end{table}

As shown in Table~\ref{tab:arpg_comparison}, ARPG achieves consistently high MSR across all benchmark tasks, indicating reliable construction of valid and task-aligned MDP formulations.
It also obtains the highest PGSR across all tasks, showing that the generated formulations can be grounded into executable environments and trained into task-satisfying policies.
Rollout videos are provided in the supplementary material, and Appendix~\ref{sec:appendix_difficulty} further analyzes where implementation corrections arise across tasks.

Gemini-3 and GPT-5.2, the two \rev{one-shot} monolithic LLM baselines, exhibit a similar failure pattern.
Although they sometimes satisfy parts of the modeling criteria, they fail to generate executable Isaac Lab environments even for the simplest CartPole task, resulting in zero EISR and therefore zero PGSR across all tasks.
\rev{Iterative GPT-5.2 extends the one-shot baseline by allowing the same model to revise its generated code using execution error logs.
It achieves full success across all three metrics on CartPole and improves EISR across all tasks.
On the other four tasks, however, reported MSR remains unchanged from one-shot GPT-5.2, while PGSR remains low.}
\rev{Thus, the one-shot monolithic baselines face bottlenecks in both MDP modeling and simulator-specific implementation. Iterative GPT-5.2 partially alleviates the implementation bottleneck through execution feedback, while task-aligned MDP formulation remains a challenge on the more complex tasks.}

\rev{A similar gap between environment executability and policy generation success is observed for Claude and Codex, the two off-the-shelf agentic systems.}
\rev{Despite this limitation, both systems obtain higher average PGSR than iterative GPT-5.2.
Their native planning and context management, together with component-wise implementation, may improve cross-module consistency and task alignment.
For example, on Reach, Codex achieves higher PGSR despite lower EISR than iterative GPT-5.2.}
\rev{However, the generated MDPs still frequently fail} the validity and task-alignment criteria captured by MSR \rev{on the more complex tasks}.
\rev{Appendix~\ref{app:reward_failure_cases} presents representative failures involving misaligned reward and termination logic, over-specified reward functions, and poorly scaled auxiliary terms that suppress the main task signal during training.}
ARPG mitigates this failure mode by externalizing dependency-aware MDP artifacts, verifying them before implementation, and applying artifact-localized correction when inconsistencies are detected.

{\color{revisionblue}
\subsection{Ablation Study}
\label{sec:ablation_study}


\begin{wraptable}{r}{0.5\textwidth}
\color{revisionblue}
\centering
\vspace{-0.5em}
\captionsetup{justification=centering}
\caption{\rev{Ablation results.}}
\label{tab:arpg_ablations}
\small
\setlength{\tabcolsep}{2pt}
\begin{tabular}{lccc}
\toprule
Task & ARPG & w/o EC & w/o sequential \\
\midrule
CartPole
& \textbf{1.0} / \textbf{1.0} / \textbf{1.0}
& \textbf{1.0} / \textbf{1.0} / \textbf{1.0}
& 0.9 / 0.9 / 0.9 \\
Ant
& \textbf{1.0} / \textbf{0.9} / \textbf{0.9}
& 0.9 / \textbf{0.9} / 0.6
& 0.3 / \textbf{0.9} / 0.1 \\
Humanoid
& \textbf{0.9} / 0.7 / \textbf{0.4}
& 0.3 / \textbf{1.0} / 0.3
& 0.1 / \textbf{1.0} / 0.1 \\
Reach
& \textbf{0.9} / 0.9 / \textbf{0.9}
& 0.5 / 0.8 / 0.1
& 0.0 / \textbf{1.0} / 0.0 \\
Lift
& \textbf{1.0} / 0.9 / \textbf{0.4}
& 0.2 / \textbf{1.0} / 0.2
& 0.0 / 0.9 / 0.0 \\
\bottomrule
\end{tabular}
\vspace{-0.6\baselineskip}
\end{wraptable}

To assess the contributions of modeling-phase EC and sequential formulation to ARPG's performance gains over the baselines, we evaluate two ablated variants.
ARPG w/o EC disables modeling-phase EC, whereas
ARPG w/o sequential generates artifacts in parallel with EC retained.
Both variants retain the same implementation phase, including
debugging through execution feedback.
All other experimental settings remain the same as in Section~\ref{sec:overall}.

In Table~\ref{tab:arpg_ablations}, both ablated variants retain high
success rates on the simpler CartPole task, indicating that removing
either component has little effect in this task.
However, both variants yield lower MSR and PGSR than ARPG on the
more complex tasks while maintaining high EISR.
In some cases, EISR even exceeds that of ARPG despite the declines
in formulation correctness and policy success.
This shared pattern shows that high environment executability does
not necessarily translate into a task-aligned MDP formulation or
a successful learned policy.

The declines in MSR and PGSR are larger for ARPG w/o sequential
than for ARPG w/o EC across these tasks.
In particular, Reach and Lift obtain zero MSR and PGSR when artifacts
are generated in parallel, despite retaining modeling-phase EC.
This pattern is consistent with the correction-interference account
in Section~\ref{subsec:theoretical_justification}, suggesting that
the effectiveness of localized correction depends on how dependencies
among formulation artifacts are organized.
Together, these results support the contributions of both components to
policy generation, with the larger decline under parallel formulation
underscoring the central role of dependency-ordered sequential formulation.
}

\subsection{Additional Generality Demonstrations Beyond Benchmark Tasks}
\label{sec:generality_demo}

\begin{wraptable}{r}{0.48\textwidth}
\centering
\vspace{-0.5em}
\caption{Examples of demonstrations.}
\label{tab:generality_main}
\begin{small}
\setlength{\tabcolsep}{2pt}
\begin{tabular}{lll}
\toprule
Setting & Examples & Backend \\
\midrule
\makecell[l]{Robotics\\variants} &
\makecell[l]{Jumping Ant,\\Jumping Humanoid,\\Pick-and-place} &
Isaac Sim \\
\midrule
Classic control &
\makecell[l]{MountainCar,\\Drone delivery} &
\makecell[l]{Gymnasium,\\Grid env.} \\
\midrule
\makecell[l]{Engineering\\MDPs} &
\makecell[l]{Wireless,\\Energy} &
Custom Env. \\
\bottomrule
\end{tabular}
\end{small}
\end{wraptable}

Beyond the five main Isaac Lab tasks, we further examine whether ARPG can be applied to tasks with different objectives, embodiments, and implementation backends.
These additional studies include custom Isaac Sim robotics tasks, standard Gymnasium examples, grid-based decision problems, and domain-specific engineering environments as summarized in Table \ref{tab:generality_main}.
These cases are not treated as success-rate benchmarks; instead, they test whether the same dependency-aware modeling pipeline can be reused across various task families while only adapting the environment-specific implementation interface.
Across these demonstrations, ARPG automatically generates coherent MDP artifacts and executable policies. 
Details of these demonstrations, including objectives and supporting evidence, are provided in Appendix~\ref{sec:additional_generality} and the supplementary material.

\subsection{Computational Cost and System Overhead}
\label{sec:cost}

We next compare the computational overhead of ARPG with off-the-shelf agentic systems.
This comparison is important because ARPG invokes multiple specialized agents and may perform localized re-invocation during verification or debugging, which could increase cost despite improving reliability.
We therefore report three quantities: direct token usage, cached token usage, and wall-clock runtime.
Direct tokens measure the explicit input/output tokens consumed during generation, cached tokens capture repeated context reuse in agentic systems, and runtime measures the end-to-end time required to produce the final pipeline.
Table~\ref{tab:computational_cost} summarizes these costs across benchmark tasks.

\begin{table}[ht]
\centering
\vspace{-0.5em}
\caption{Computational cost comparison. Token usage is reported as total tokens, with direct and cached tokens shown in parentheses. Bold indicates the lowest value within each task.}
\label{tab:computational_cost}
\begin{small}
\setlength{\tabcolsep}{4pt}
\begin{tabular}{lcccccc}
\toprule
\multirow{2}{*}{Task}
& \multicolumn{3}{c}{Tokens (M)}
& \multicolumn{3}{c}{Runtime (min)} \\
\cmidrule(lr){2-4}
\cmidrule(lr){5-7}
& ARPG & Claude & Codex
& ARPG & Claude & Codex \\
\midrule

CartPole
& \textbf{0.160} (0.140, 0.020)
& 2.586 (0.036, 2.550)
& 2.665 (0.199, 2.466)
& \textbf{2.9}
& 8.2
& 7.4 \\

Ant
& \textbf{0.149} (0.129, 0.020)
& 5.711 (0.061, 5.650)
& 2.671 (0.177, 2.494)
& \textbf{3.9}
& 17.2
& 7.7 \\

Humanoid
& \textbf{0.248} (0.208, 0.040)
& 7.882 (0.092, 7.790)
& 4.201 (0.244, 3.957)
& \textbf{4.5}
& 24.1
& 11.8 \\

Reach
& \textbf{0.195} (0.165, 0.030)
& 1.818 (0.178, 1.640)
& 2.433 (0.173, 2.260)
& \textbf{4.6}
& 11.8
& 9.3 \\

Lift
& \textbf{0.232} (0.192, 0.040)
& 1.769 (0.119, 1.650)
& 2.315 (0.165, 2.150)
& \textbf{6.2}
& 15.0
& 8.2 \\

\bottomrule
\end{tabular}
\end{small}
\end{table}

Although ARPG invokes multiple specialized agents, \rev{it incurs lower measured computational overhead than the evaluated off-the-shelf systems}.
Each ARPG agent receives a bounded role-specific context and operates on compact intermediate artifacts, whereas off-the-shelf systems may repeatedly reintroduce full documents and accumulated interaction histories during code generation and debugging.
\rev{In our evaluation}, ARPG uses fewer \rev{reported} total tokens and \rev{has a} shorter wall-clock runtime across all tasks.
The gap is largest on Humanoid: ARPG takes 4.5 minutes with 0.248 M tokens, while Claude and Codex require 24.1 and 11.8 minutes, respectively, with 7.882 and 4.201 M tokens.
\rev{These results show a practical efficiency advantage for ARPG under the evaluated configurations, with each system retaining its native scaffolding and token accounting.}




\section{Conclusion}

In this paper, we studied automated generation of executable robot RL policies from natural-language task descriptions.
We proposed ARPG, an agentic framework that treats LLMs as MDP designers and enforces staged formulation, artifact externalization, verification, and localized correction.
Experiments on Isaac Lab control, locomotion, and manipulation tasks show that ARPG improves modeling correctness and policy-generation success over monolithic LLM baselines and off-the-shelf agentic systems.
The results also indicate that executable implementation alone is insufficient: reliable policy generation requires dependency-aligned state, action, reward, and termination artifacts.
Additional demonstrations on custom robotics, classic controls, and engineering-domain MDPs suggest that ARPG can be instantiated beyond the main benchmark suite.
\clearpage
{\color{revisionblue}
\begin{ack}
This work was supported in part by the National Research Foundation of Korea (NRF) grant funded by the Korea government (MSIT) (RS-2025-24523498 \& RS-2026-25482257),
in part by the IITP (Institute of Information \& Communications Technology Planning \& Evaluation)--ITRC (Information Technology Research Center) grant funded by the Korea government (Ministry of Science and ICT) (IITP-2026-RS-2021-II211816),
and in part by the Technology Innovation Program (RS-2022-00154678, Development of Intelligent Sensor Platform Technology for Connected Sensor) funded by the Ministry of Trade, Industry \& Energy (MOTIE, Korea).
We thank all reviewers for their valuable comments and suggestions. The authors declare no financial conflicts of interest related to this work.
\end{ack}
}

\bibliographystyle{unsrtnat}
\bibliography{example_paper}

\begin{thebibliography}{35}
\providecommand{\natexlab}[1]{#1}
\providecommand{\url}[1]{\texttt{#1}}
\expandafter\ifx\csname urlstyle\endcsname\relax
  \providecommand{\doi}[1]{doi: #1}\else
  \providecommand{\doi}{doi: \begingroup \urlstyle{rm}\Url}\fi

\bibitem[Zeng et~al.(2023)Zeng, Gan, Wang, Liu, and Yu]{zeng2023large}
Fanlong Zeng, Wensheng Gan, Yongheng Wang, Ning Liu, and Philip~S Yu.
\newblock Large language models for robotics: A survey.
\newblock \emph{arXiv preprint arXiv:2311.07226}, 2023.

\bibitem[Kawaharazuka et~al.(2025)Kawaharazuka, Oh, Yamada, Posner, and
  Zhu]{kawaharazuka2025vision}
Kento Kawaharazuka, Jihoon Oh, Jun Yamada, Ingmar Posner, and Yuke Zhu.
\newblock Vision-language-action models for robotics: A review towards
  real-world applications.
\newblock \emph{IEEE Access}, 2025.

\bibitem[Wang et~al.(2025)Wang, Shi, Hu, Ma, Liu, Wang, Yao, Liu, Ge, and
  Zhang]{wang2025large}
Jiaqi Wang, Enze Shi, Huawen Hu, Chong Ma, Yiheng Liu, Xuhui Wang, Yincheng
  Yao, Xuan Liu, Bao Ge, and Shu Zhang.
\newblock Large language models for robotics: Opportunities, challenges, and
  perspectives.
\newblock \emph{Journal of Automation and Intelligence}, 4\penalty0
  (1):\penalty0 52--64, 2025.

\bibitem[Liu et~al.(2025)Liu, Sun, and Kapadia]{liu2025integrating}
Yutong Liu, Qingquan Sun, and Dhruvi~Rajeshkumar Kapadia.
\newblock Integrating large language models into robotic autonomy: A review of
  motion, voice, and training pipelines.
\newblock \emph{AI}, 6\penalty0 (7):\penalty0 158, 2025.

\bibitem[Brunke et~al.(2022)Brunke, Greeff, Hall, Yuan, Zhou, Panerati, and
  Schoellig]{brunke2022safe}
Lukas Brunke, Melissa Greeff, Adam~W Hall, Zhaocong Yuan, Siqi Zhou, Jacopo
  Panerati, and Angela~P Schoellig.
\newblock Safe learning in robotics: From learning-based control to safe
  reinforcement learning.
\newblock \emph{Annual Review of Control, Robotics, and Autonomous Systems},
  5\penalty0 (1):\penalty0 411--444, 2022.

\bibitem[Levine et~al.(2018)Levine, Pastor, Krizhevsky, Ibarz, and
  Quillen]{levine2018learning}
Sergey Levine, Peter Pastor, Alex Krizhevsky, Julian Ibarz, and Deirdre
  Quillen.
\newblock Learning hand-eye coordination for robotic grasping with deep
  learning and large-scale data collection.
\newblock \emph{The International Journal of Robotics Research}, 37\penalty0
  (4-5):\penalty0 421--436, 2018.

\bibitem[Peng et~al.(2020)Peng, Coumans, Zhang, Lee, Tan, and
  Levine]{peng2020learning}
Xue~Bin Peng, Erwin Coumans, Tingnan Zhang, Tsang-Wei Lee, Jie Tan, and Sergey
  Levine.
\newblock Learning agile robotic locomotion skills by imitating animals.
\newblock \emph{arXiv preprint arXiv:2004.00784}, 2020.

\bibitem[Kober et~al.(2013)Kober, Bagnell, and Peters]{kober2013reinforcement}
Jens Kober, J~Andrew Bagnell, and Jan Peters.
\newblock Reinforcement learning in robotics: A survey.
\newblock \emph{The International Journal of Robotics Research}, 32\penalty0
  (11):\penalty0 1238--1274, 2013.

\bibitem[Dulac-Arnold et~al.(2021)Dulac-Arnold, Levine, Mankowitz, Li,
  Paduraru, Gowal, and Hester]{dulac2021challenges}
Gabriel Dulac-Arnold, Nir Levine, Daniel~J Mankowitz, Jerry Li, Cosmin
  Paduraru, Sven Gowal, and Todd Hester.
\newblock Challenges of real-world reinforcement learning: definitions,
  benchmarks and analysis.
\newblock \emph{Machine Learning}, 110\penalty0 (9):\penalty0 2419--2468, 2021.

\bibitem[Henderson et~al.(2018)Henderson, Islam, Bachman, Pineau, Precup, and
  Meger]{henderson2018deep}
Peter Henderson, Riashat Islam, Philip Bachman, Joelle Pineau, Doina Precup,
  and David Meger.
\newblock Deep reinforcement learning that matters.
\newblock In \emph{Proceedings of the AAAI Conference on Artificial
  Intelligence}, volume~32, 2018.

\bibitem[Andrychowicz et~al.(2020)Andrychowicz, Baker, Chociej, Jozefowicz,
  McGrew, Pachocki, Petron, Plappert, Powell, Ray,
  et~al.]{andrychowicz2020learning}
OpenAI:~Marcin Andrychowicz, Bowen Baker, Maciek Chociej, Rafal Jozefowicz, Bob
  McGrew, Jakub Pachocki, Arthur Petron, Matthias Plappert, Glenn Powell, Alex
  Ray, et~al.
\newblock Learning dexterous in-hand manipulation.
\newblock \emph{The International Journal of Robotics Research}, 39\penalty0
  (1):\penalty0 3--20, 2020.

\bibitem[Wei et~al.(2022)Wei, Wang, Schuurmans, Bosma, Xia, Chi, Le, Zhou,
  et~al.]{wei2022chain}
Jason Wei, Xuezhi Wang, Dale Schuurmans, Maarten Bosma, Fei Xia, Ed~Chi, Quoc~V
  Le, Denny Zhou, et~al.
\newblock Chain-of-thought prompting elicits reasoning in large language
  models.
\newblock \emph{Advances in Neural Information Processing Systems},
  35:\penalty0 24824--24837, 2022.

\bibitem[Chen(2021)]{chen2021evaluating}
Mark Chen.
\newblock Evaluating large language models trained on code.
\newblock \emph{arXiv preprint arXiv:2107.03374}, 2021.

\bibitem[Yao et~al.(2022)Yao, Zhao, Yu, Du, Shafran, Narasimhan, and
  Cao]{yao2022react}
Shunyu Yao, Jeffrey Zhao, Dian Yu, Nan Du, Izhak Shafran, Karthik~R Narasimhan,
  and Yuan Cao.
\newblock Re{A}ct: Synergizing reasoning and acting in language models.
\newblock In \emph{International Conference on Learning Representations}, 2022.

\bibitem[Driess et~al.(2023)Driess, Xia, Sajjadi, Lynch, Chowdhery, Ichter,
  Wahid, Tompson, Vuong, Yu, et~al.]{driess2023palm}
Danny Driess, Fei Xia, Mehdi~SM Sajjadi, Corey Lynch, Aakanksha Chowdhery,
  Brian Ichter, Ayzaan Wahid, Jonathan Tompson, Quan Vuong, Tianhe Yu, et~al.
\newblock {PaLM-E}: an embodied multimodal language model.
\newblock In \emph{Proceedings of the International Conference on Machine
  Learning}, pages 8469--8488, 2023.

\bibitem[Brohan et~al.(2022)Brohan, Brown, Carbajal, Chebotar, Dabis, Finn,
  Gopalakrishnan, Hausman, Herzog, Hsu, et~al.]{brohan2022rt}
Anthony Brohan, Noah Brown, Justice Carbajal, Yevgen Chebotar, Joseph Dabis,
  Chelsea Finn, Keerthana Gopalakrishnan, Karol Hausman, Alex Herzog, Jasmine
  Hsu, et~al.
\newblock {RT-1}: Robotics transformer for real-world control at scale.
\newblock \emph{arXiv preprint arXiv:2212.06817}, 2022.

\bibitem[O’Neill et~al.(2024)O’Neill, Rehman, Maddukuri, Gupta, Padalkar,
  Lee, Pooley, Gupta, Mandlekar, Jain, et~al.]{o2024open}
Abby O’Neill, Abdul Rehman, Abhiram Maddukuri, Abhishek Gupta, Abhishek
  Padalkar, Abraham Lee, Acorn Pooley, Agrim Gupta, Ajay Mandlekar, Ajinkya
  Jain, et~al.
\newblock Open x-embodiment: Robotic learning datasets and rt-x models: Open
  x-embodiment collaboration 0.
\newblock In \emph{Proceedings of the IEEE International Conference on Robotics
  and Automation}, pages 6892--6903. IEEE, 2024.

\bibitem[Reed et~al.(2022)Reed, Zolna, Parisotto, Colmenarejo, Novikov,
  Barth-Maron, Gimenez, Sulsky, Kay, Springenberg, et~al.]{reed2022generalist}
Scott Reed, Konrad Zolna, Emilio Parisotto, Sergio~Gomez Colmenarejo, Alexander
  Novikov, Gabriel Barth-Maron, Mai Gimenez, Yury Sulsky, Jackie Kay,
  Jost~Tobias Springenberg, et~al.
\newblock A generalist agent.
\newblock \emph{arXiv preprint arXiv:2205.06175}, 2022.

\bibitem[Zitkovich et~al.(2023)Zitkovich, Yu, Xu, Xu, Xiao, Xia, Wu, Wohlhart,
  Welker, Wahid, et~al.]{zitkovich2023rt}
Brianna Zitkovich, Tianhe Yu, Sichun Xu, Peng Xu, Ted Xiao, Fei Xia, Jialin Wu,
  Paul Wohlhart, Stefan Welker, Ayzaan Wahid, et~al.
\newblock {RT-2}: Vision-language-action models transfer web knowledge to
  robotic control.
\newblock In \emph{Proceedings of the Conference on Robot Learning}, pages
  2165--2183. PMLR, 2023.

\bibitem[Kim et~al.(2024)Kim, Pertsch, Karamcheti, Xiao, Balakrishna, Nair,
  Rafailov, Foster, Lam, Sanketi, et~al.]{kim2024openvla}
Moo~Jin Kim, Karl Pertsch, Siddharth Karamcheti, Ted Xiao, Ashwin Balakrishna,
  Suraj Nair, Rafael Rafailov, Ethan Foster, Grace Lam, Pannag Sanketi, et~al.
\newblock Open{VLA}: An open-source vision-language-action model.
\newblock \emph{arXiv preprint arXiv:2406.09246}, 2024.

\bibitem[Ter et~al.(2025)Ter, Ajayi, and Udekwe]{ter2025taxonomy}
Kumater Ter, Ore-Ofe Ajayi, and Daniel Udekwe.
\newblock Taxonomy and trends in reinforcement learning for robotics and
  control systems: A structured review.
\newblock \emph{arXiv preprint arXiv:2510.21758}, 2025.

\bibitem[Kaufmann et~al.(2024)Kaufmann, Weng, Bengs, and
  H{\"u}llermeier]{kaufmann2024survey}
Timo Kaufmann, Paul Weng, Viktor Bengs, and Eyke H{\"u}llermeier.
\newblock A survey of reinforcement learning from human feedback.
\newblock \emph{Transactions on Machine Learning Research}, 2024.
\newblock ISSN 2835-8856.

\bibitem[Chakraborty et~al.(2023)Chakraborty, Singh, Bhaskar, Tokekar, Manocha,
  and Bedi]{chakraborty2023rebel}
Souradip Chakraborty, Anukriti Singh, Amisha Bhaskar, Pratap Tokekar, Dinesh
  Manocha, and Amrit~Singh Bedi.
\newblock {REBEL}: A regularization-based solution for reward overoptimization
  in robotic reinforcement learning from human feedback.
\newblock \emph{arXiv preprint arXiv:2312.14436}, 2023.

\bibitem[Ma et~al.(2023)Ma, Liang, Wang, Huang, Bastani, Jayaraman, Zhu, Fan,
  and Anandkumar]{ma2023eureka}
Yecheng~Jason Ma, William Liang, Guanzhi Wang, De-An Huang, Osbert Bastani,
  Dinesh Jayaraman, Yuke Zhu, Linxi Fan, and Anima Anandkumar.
\newblock Eureka: Human-level reward design via coding large language models.
\newblock \emph{arXiv preprint arXiv:2310.12931}, 2023.

\bibitem[Lee et~al.(2026)Lee, Wagenmaker, Pertsch, Liang, Levine, and
  Finn]{lee2026roboreward}
Tony Lee, Andrew Wagenmaker, Karl Pertsch, Percy Liang, Sergey Levine, and
  Chelsea Finn.
\newblock Roboreward: General-purpose vision-language reward models for
  robotics.
\newblock \emph{arXiv preprint arXiv:2601.00675}, 2026.

\bibitem[Schick et~al.(2023)Schick, Dwivedi-Yu, Dess{\`\i}, Raileanu, Lomeli,
  Hambro, Zettlemoyer, Cancedda, and Scialom]{schick2023toolformer}
Timo Schick, Jane Dwivedi-Yu, Roberto Dess{\`\i}, Roberta Raileanu, Maria
  Lomeli, Eric Hambro, Luke Zettlemoyer, Nicola Cancedda, and Thomas Scialom.
\newblock Toolformer: Language models can teach themselves to use tools.
\newblock \emph{Advances in Neural Information Processing Systems},
  36:\penalty0 68539--68551, 2023.

\bibitem[Yao et~al.(2023)Yao, Yu, Zhao, Shafran, Griffiths, Cao, and
  Narasimhan]{yao2023tree}
Shunyu Yao, Dian Yu, Jeffrey Zhao, Izhak Shafran, Tom Griffiths, Yuan Cao, and
  Karthik Narasimhan.
\newblock Tree of thoughts: Deliberate problem solving with large language
  models.
\newblock \emph{Advances in Neural Information Processing Systems},
  36:\penalty0 11809--11822, 2023.

\bibitem[Shinn et~al.(2023)Shinn, Cassano, Gopinath, Narasimhan, and
  Yao]{shinn2023reflexion}
Noah Shinn, Federico Cassano, Ashwin Gopinath, Karthik Narasimhan, and Shunyu
  Yao.
\newblock Reflexion: Language agents with verbal reinforcement learning.
\newblock \emph{Advances in Neural Information Processing Systems},
  36:\penalty0 8634--8652, 2023.

\bibitem[Madaan et~al.(2023)Madaan, Tandon, Gupta, Hallinan, Gao, Wiegreffe,
  Alon, Dziri, Prabhumoye, Yang, et~al.]{madaan2023self}
Aman Madaan, Niket Tandon, Prakhar Gupta, Skyler Hallinan, Luyu Gao, Sarah
  Wiegreffe, Uri Alon, Nouha Dziri, Shrimai Prabhumoye, Yiming Yang, et~al.
\newblock Self-refine: Iterative refinement with self-feedback.
\newblock \emph{Advances in Neural Information Processing Systems},
  36:\penalty0 46534--46594, 2023.

\bibitem[Le et~al.(2022)Le, Wang, Gotmare, Savarese, and Hoi]{le2022coderl}
Hung Le, Yue Wang, Akhilesh~Deepak Gotmare, Silvio Savarese, and Steven
  Chu~Hong Hoi.
\newblock Coderl: Mastering code generation through pretrained models and deep
  reinforcement learning.
\newblock \emph{Advances in Neural Information Processing Systems},
  35:\penalty0 21314--21328, 2022.

\bibitem[Yang et~al.(2023)Yang, Wang, Lu, Liu, Le, Zhou, and
  Chen]{yang2023large}
Chengrun Yang, Xuezhi Wang, Yifeng Lu, Hanxiao Liu, Quoc~V Le, Denny Zhou, and
  Xinyun Chen.
\newblock Large language models as optimizers.
\newblock In \emph{International Conference on Learning Representations}, 2023.

\bibitem[Astorga et~al.(2025)Astorga, Liu, Xiao, and van~der
  Schaar]{astorga2024autoformulation}
Nicol{\'a}s Astorga, Tennison Liu, Yuanzhang Xiao, and Mihaela van~der Schaar.
\newblock Autoformulation of mathematical optimization models using {LLMs}.
\newblock In \emph{Proceedings of the International Conference on Machine
  Learning}, 2025.

\bibitem[AhmadiTeshnizi et~al.(2024)AhmadiTeshnizi, Gao, Brunborg, Talaei,
  Lawless, and Udell]{ahmaditeshnizi2024optimus}
Ali AhmadiTeshnizi, Wenzhi Gao, Herman Brunborg, Shayan Talaei, Connor Lawless,
  and Madeleine Udell.
\newblock {OptiMUS-0.3}: Using large language models to model and solve
  optimization problems at scale.
\newblock \emph{arXiv preprint arXiv:2407.19633}, 2024.

\bibitem[Kim et~al.(2025)Kim, Gu, Park, Park, Schmidgall, Heydari, Yan, Zhang,
  Zhuang, Malhotra, et~al.]{kim2025towards}
Yubin Kim, Ken Gu, Chanwoo Park, Chunjong Park, Samuel Schmidgall, A~Ali
  Heydari, Yao Yan, Zhihan Zhang, Yuchen Zhuang, Mark Malhotra, et~al.
\newblock Towards a science of scaling agent systems.
\newblock \emph{arXiv preprint arXiv:2512.08296}, 2025.

\bibitem[Lee et~al.(2025)Lee, Kim, Lee, Moon, Hwang, Kim, Neubig, Welleck, and
  Choi]{leerefinebench}
Young-Jun Lee, Seungone Kim, Byung-Kwan Lee, Minkyeong Moon, Yechan Hwang,
  Jong~Myoung Kim, Graham Neubig, Sean Welleck, and Ho-Jin Choi.
\newblock {RefineBench}: Evaluating refinement capability in language models.
\newblock In \emph{First Workshop on Multi-Turn Interactions in Large Language
  Models}, 2025.

\end{thebibliography}
\clearpage
\appendix

\section{Discussion and Future Directions}
\label{sec:discussion}

From our experimental analysis, we identified several significant limitations of the ARPG framework, as well as opportunities to extend it.
\paragraph{Parameter Sensitivity in Policy Training.}
Even after a semantically valid MDP formulation and executable environment are obtained, some failures still occur during policy training.
These failures often arise from numerical calibration issues, such as improperly balanced reward weights or coordinate-frame inconsistencies, rather than from missing state, action, reward, or termination semantics.
These issues do not reflect fundamental limitations of the proposed framework, but rather stem from its sensitivity to implementation-level parameters that are only weakly constrained by natural language descriptions.
As experience accumulates across tasks and environments, such parameterization errors could be progressively mitigated through data-driven calibration, meta-optimization of reward coefficients, or improved priors over common coordinate conventions.

\paragraph{Trade-offs in Agent Granularity.}
We observed that the ARPG framework achieves high success rates, partly due to its relatively fine-grained decomposition of the formulation process.
However, increased agent granularity can incur higher computational and budget costs when agent contexts and interactions are not explicitly bounded.
In ARPG, role-specific contexts and compact artifact interfaces mitigate this overhead, which is consistent with the lower token usage and shorter runtime observed against general-purpose agentic systems.
These results suggest that a more fine-grained decomposition is not universally optimal, but instead reflects a trade-off between formulation success, inference cost, and the level of reasoning transparency. Depending on the complexity of the task and
the desired degree of interpretability, different levels of agent granularity may be preferable.

Importantly, the optimal degree of decomposition is also expected to evolve with advances in single-agent reasoning capabilities. As individual language models become more reliable at handling longer dependency chains and structured verification internally, fewer formulation agents may suffice to achieve comparable performance.
This motivates the need for a manager-level agent that can adaptively control the
degree of agent decomposition based on the task description, target transparency,
resource constraints, and the underlying capabilities of the language model, rather than relying on a fixed agent structure.

\paragraph{Simulator-Specific Structure in Environment Implementation.}
During the implementation phase, variations in configuration structures that depend on tasks and simulators pose a recurring challenge. Different simulators expose distinct component abstractions, dependency structures, and configuration units. Even within the same simulator, different tasks may require different combinations of managers (e.g., scene, observation, command, reward, and termination). Incorrect assumptions about these structures often lead to subtle runtime errors.
While grounding code generation in simulator documentation via retrieval-augmented generation (RAG) can improve robustness within a fixed simulator, a more general approach is to design an agent that explicitly reasons about the target simulator architecture itself. Such an agent could dynamically determine how the implementation phase should be decomposed, selecting the appropriate configuration modules, agent roles, and generation order, based on the simulator's structural properties.
This would enable the framework to adapt its implementation-agent structure across different simulators, moving toward a simulator-agnostic policy generation framework rather than one tailored to a single execution backend.

\paragraph{Prompt Engineering and Reuse.}
Although the prompts and per-configuration documents used by ARPG may appear lengthy, most of their content specifies reusable simulator-interface structure rather than task-specific tuning.
Within Isaac Lab, the same agent prompts are reused across tasks without modification; task variation is introduced through the natural language task description, robot specification, and the corresponding configuration document.
Thus, the prompt design is primarily a reusable interface specification for the target environment, rather than per-task prompt engineering.

\paragraph{Generality Beyond Robotics.}
Although this paper focuses on robotic policy generation, ARPG is not conceptually tied to robot morphology or to Isaac Lab.
Its core assumption is that a task can be formulated as an MDP through dependent choices over states, actions, objectives, constraints, rewards, and termination conditions.
This structure also appears in engineering domains such as wireless resource allocation and energy management, where task objectives and constraints must be reformulated under changing operating conditions.
The additional demonstrations in Appendix~\ref{sec:additional_generality} suggest that the dependency-aware modeling stage can transfer beyond the main robotics benchmark suite.
However, each new execution backend still requires an implementation interface that exposes the relevant environment dynamics, constraints, and executable code structure.
\rev{The evaluated scope is state-based MDPs with supported simulator interfaces; visual perception and long-horizon compositional task planning are not demonstrated.}


\paragraph{Toward Integrated Autonomous Policy Generation.}
More broadly, these findings highlight that end-to-end automated RL policy generation is a multi-faceted problem, spanning formulation, environment implementation, and policy training.
While ARPG primarily addresses dependency-aware formulation and localized implementation correction, future systems will benefit from integrating adaptive decomposition strategies, documentation- and architecture-aware implementation agents, and training-time optimization mechanisms.
We believe that these extensions are a promising step toward fully autonomous and scalable robot policy generation frameworks.

\paragraph{Broader Impacts.}
\label{app:broader_impact}
ARPG is intended to reduce the manual effort required to formulate and implement robot RL policies.
Because the framework operates before deployment, the generated policies are standalone and do not rely on language models during execution, avoiding runtime LLM decision-making in the control loop.
However, this does not provide formal safety guarantees for the generated policies.
As with other robotics automation tools, the framework may improve development efficiency and accessibility while raising broader concerns related to labor displacement or misuse.
We do not identify immediate negative societal impacts unique to this work beyond those commonly associated with robotic automation and RL-based control.

\section{Theoretical Analysis on Correction Dynamics and Modeling Success}
\label{app:correction_dynamics}

This appendix provides a \rev{conditional explanatory analysis} of the correction dynamics underlying
Proposition~\ref{prop:relative_improvement}.
We model formulation as an iterative verification-correction process over explicit formulation artifacts.
Let
\[
p_i^{(t)} := \mathbb{P}(\text{artifact } i \text{ is invalid after iteration } t).
\]
As in Section~\ref{subsec:theoretical_justification}, this artifact-wise probability is interpreted conditionally on the validity of preceding artifacts in the sequential formulation chain. The product form in Equation~\eqref{eq:sm_def} therefore follows the chain rule for successful sequential construction, rather than requiring independent artifact errors.

We characterize the correction dynamics using two effective parameters:
\begin{enumerate}
\item[(i)] \textbf{Effective correctability} $\alpha_i > 0$, the probability that
an invalid artifact is successfully corrected in a single iteration.
\item[(ii)]\textbf{Error reintroduction rate} $\beta_i \in [0,1)$, the probability
that a valid artifact becomes invalid due to the correction of other artifacts.
\end{enumerate}
Under these assumptions, the artifact-wise error probability satisfies
\begin{equation}
\label{eq:recurrence}
p_i^{(t+1)}
\le
(1 - \alpha_i - \beta_i)\, p_i^{(t)} + \beta_i .
\end{equation}

Equation~\eqref{eq:recurrence} defines a stable linear recurrence.
For each formulation artifact $i$, the residual error probability satisfies
\begin{equation}
\label{eq:residual_bound}
p_i^{(\infty)}
:=
\limsup_{t \to \infty} p_i^{(t)}
\le
\frac{\beta_i}{\alpha_i + \beta_i}.
\end{equation}
We note that the residual error bound in
Equation~\eqref{eq:residual_bound} exhibits monotonicity with respect to the
effective correction dynamics parameters.
Specifically, for each formulation artifact $i$,
\[
\frac{\partial}{\partial \alpha_i}
\left(
\frac{\beta_i}{\alpha_i + \beta_i}
\right)
=
-\frac{\beta_i}{(\alpha_i + \beta_i)^2}
\mathrel{\rev{\le}} 0,
\]
and
\[
\frac{\partial}{\partial \beta_i}
\left(
\frac{\beta_i}{\alpha_i + \beta_i}
\right)
=
\frac{\alpha_i}{(\alpha_i + \beta_i)^2}
> 0.
\]
Thus, increasing effective correctability $\alpha_i$ \rev{does not increase} the
residual error \rev{bound}, while increasing error reintroduction $\beta_i$
strictly increases it.

\paragraph{Proof of Proposition~\ref{prop:relative_improvement}.}

Consider two formulation strategies, ARPG and a baseline, with residual error probabilities $p_i^{(\infty,\mathrm{ARPG})}$ and
$p_i^{(\infty,\mathrm{base})}$, respectively.
Suppose that for all $i$,
\[
p_i^{(\infty,\mathrm{ARPG})}
\le
p_i^{(\infty,\mathrm{base})},
\]
with strict inequality for at least one artifact $k$.
Then, by Equation~\eqref{eq:sm_def} under the conditional residual-error interpretation,
\[
\mathbb{P}(S_M^{\mathrm{ARPG}})
=
\prod_i \bigl(1 - p_i^{(\infty,\mathrm{ARPG})}\bigr)
>
\prod_i \bigl(1 - p_i^{(\infty,\mathrm{base})}\bigr)
=
\mathbb{P}(S_M^{\mathrm{base}}).
\]

Equation~\eqref{eq:residual_bound} implies that any formulation strategy that
induces higher effective correctability $\alpha_i$ or lower error reintroduction
$\beta_i$ yields a \rev{weakly smaller residual error bound}.
Therefore, under the \rev{condition} that ARPG \rev{reduces at least one residual
artifact error probability without increasing the others} relative to a baseline,
Proposition~\ref{prop:relative_improvement} follows.

\paragraph{Relation to Single-Agent and Parallel Strategies.}

A single-agent formulation strategy that generates a formulation in a single
pass \rev{without iterative correction yields}
$p_i^{(\infty)} = p_i^{(0)}$.
Parallel formulation strategies that modify multiple artifacts simultaneously
\rev{may increase} cross-artifact interference, corresponding to higher effective
$\beta_i$.
\rev{Under the assumed correction dynamics, these architectural differences may lead to larger residual errors and lower modeling success.}
These cases illustrate how architectural choices \rev{may influence}
the effective correction dynamics captured by the above analysis.

\section{Failure Cases of Executable but Non-learnable Environments}
\label{app:reward_failure_cases}

This appendix provides representative failure cases observed in the generated MDP and reward artifacts of the off-the-shelf agentic baselines.
The cases come from manually inspected artifacts that are associated with executable generated environments that nevertheless failed the stricter MSR/PGSR criteria.
They illustrate the distinction emphasized in Section~\ref{sec:overall}: simulator-level execution does not guarantee a learnable or semantically aligned MDP.
We group them by the artifact dependency that is broken, using concrete patterns from the inspected Ant, Humanoid, and Franka Lift artifacts.
The snippets below are shown with only minimal editorial modifications applied for presentation.
Highlighted terms indicate the artifact components most directly involved in the failure pattern.

\subsection{Reward Terms Depend on Unavailable or Inconsistent State Variables}

\paragraph{Ant frame convention mismatch.}
Some Ant MDPs define the observation using body-frame base velocity while the reward is written in terms of world-frame forward velocity along global \(+X\), yaw, heading projection, and target direction.
This is valid only if the implementation explicitly reconstructs world-frame quantities.
In the failed artifacts, that conversion is left ambiguous, so body-frame velocity can be used as if it were world-frame progress, making the forward reward inconsistent after yaw changes.

\begin{tcolorbox}[title=Observed failure snippet: Ant frame mismatch (Codex), colback=artifactred!3!white, colframe=artifactred!60!black, boxrule=0.6pt]
\footnotesize
\[
\begin{array}{ll}
\text{Observation:} & \textcolor{artifactred}{v_{\mathrm{base}}\ \text{in body/root frame}} \\
\text{Reward:} & \textcolor{artifactred}{r_{\mathrm{fwd}}=\max(v_{x,w},0),\quad
r_{\mathrm{heading}}=\mathrm{ReLU}(\cos(\mathrm{yaw}))}
\end{array}
\]
The artifact mixes body-frame observations with world-frame reward quantities without specifying the required frame conversion.
\end{tcolorbox}

\subsection{Termination and Reward Are Semantically Inconsistent}

\paragraph{Humanoid contact and height criteria.}
In Codex Humanoid artifacts, broad non-foot contacts and low body height are used both as dense penalties and as hard failure events.
One representative artifact penalizes low height below \(1.10\)m and any non-foot contact above \(1\)N, while termination fires when root height falls below \(0.85\)m or when non-foot contact exceeds \(10\)N.
Isaac Lab locomotion examples do use contact penalties and illegal-contact termination, but they typically separate the intended body set and threshold semantics.
Here, crouching, incidental non-foot contact, or recovery attempts can be treated as both heavily penalized behavior and terminal failure.
This can make recovery dynamics difficult to learn.

\begin{tcolorbox}[title=Observed failure snippet: Humanoid termination (Codex), colback=artifactred!3!white, colframe=artifactred!60!black, boxrule=0.6pt]
\footnotesize
\[
\begin{array}{ll}
\text{Reward:} &
\textcolor{artifactred}{-3.0\,\max(1.10-z_{\mathrm{root}},0)
-2.0\,N_{\mathrm{nonfoot}}(F>1\mathrm{N})} \\
\text{Termination:} &
\textcolor{artifactred}{z_{\mathrm{root}}<0.85\quad \vee\quad
N_{\mathrm{nonfoot}}(F>10\mathrm{N})>0}.
\end{array}
\]
The artifact does not clearly separate recoverable contact/height deviations from terminal fall events, making the failure boundary too broad.
\end{tcolorbox}

\paragraph{Lift drop threshold versus lift shaping.}
In several Lift artifacts, reward shaping is expressed relative to the object's initial height, while termination is specified as an absolute object-height threshold such as \(z_{\mathrm{object}}<-0.05\), \(z_{\mathrm{object}}<0.25\), or \(z_{\mathrm{object}}<0.32\), depending on the run.
An absolute threshold is valid when anchored to the same scene frame, as in the successful ARPG Lift MDP.
This also matches Isaac Lab's Lift example, which uses an absolute object-height drop threshold.
The baseline issue is narrower: some artifacts do not state how the threshold is tied to tabletop height, object size, reset distribution, or height-frame convention.
The termination rule can therefore be ambiguous relative to the shaping objective.

\begin{tcolorbox}[title=Observed failure snippet: Lift drop threshold (Claude), colback=artifactred!3!white, colframe=artifactred!60!black, boxrule=0.6pt]
\footnotesize
\[
\begin{array}{ll}
\text{Reward shaping:} &
\textcolor{artifactred}{h_{\mathrm{lift}}=z_{\mathrm{object}}-z_{\mathrm{init}}} \\
\text{Termination examples:} &
\textcolor{artifactred}{z_{\mathrm{object}}<\{-0.05,\ 0.25,\ 0.32\}\ \text{without a shared scene/frame anchor}}.
\end{array}
\]
The failure is not the use of an absolute threshold itself, but the lack of an explicit scene/frame anchor for how that threshold matches the shaping objective.
\end{tcolorbox}

\subsection{Sparse or Delayed Task Signals}

\paragraph{Lift reward gated by a hard grasp predicate.}
Many Lift artifacts reward object height only after a binary \(\mathrm{is\_grasped}\) predicate becomes true.
Before that predicate is satisfied, approach and aperture shaping may exist, but the lift and success signals are withheld.
The problem is therefore not sparse shaping in general, but delaying the main object-control signal behind a brittle grasp proxy.

\begin{tcolorbox}[title=Observed failure snippet: delayed Lift signal (Claude), colback=artifactred!3!white, colframe=artifactred!60!black, boxrule=0.6pt]
\footnotesize
\[
\begin{aligned}
r_{\mathrm{grasp}} &= \Ind[d_{\mathrm{grasp}}<0.10]\exp(-|\mathrm{aperture}-0.04|/0.025),\\
\textcolor{artifactred}{r_{\mathrm{lift}}} &= \textcolor{artifactred}{\Ind[\mathrm{is\_grasped}]}\tanh(h_{\mathrm{lift}}/0.06),\\
\textcolor{artifactred}{r_{\mathrm{success}}} &= \textcolor{artifactred}{\Ind[\mathrm{is\_grasped}]\,\Ind[h_{\mathrm{lift}}>0.04]\,2.0}.
\end{aligned}
\]
The lift and success signals are unavailable until a hard grasp predicate is already satisfied.
\end{tcolorbox}

\paragraph{Ant forward progress gated by stability.}
In some locomotion artifacts, forward reward is multiplied by uprightness and heading alignment rather than given as a standalone dense velocity signal.
Early policies that are not already upright and aligned may receive an attenuated movement reward.
The agent must therefore improve stability before forward motion becomes strongly profitable.

\begin{tcolorbox}[title=Observed failure snippet: gated locomotion signal (Claude), colback=artifactred!3!white, colframe=artifactred!60!black, boxrule=0.6pt]
\footnotesize
\[
\textcolor{artifactred}{r_{\mathrm{fwd}} = v_x^+\,r_{\mathrm{heading}}\,r_{\mathrm{upright}}}
\]
Forward progress remains present but is strongly attenuated until heading and uprightness are sufficiently good.
\end{tcolorbox}

\subsection{Auxiliary Reward Terms Obscure the Primary Task Objective}

The following cases involve reward decompositions in which auxiliary regularization terms dominate or gate the primary objective signal.
Although such failures partly involve numerical calibration, they also reflect a formulation-level dependency error: the primary objective is not preserved as a direct reward signal before auxiliary terms are added.

\paragraph{Ant locomotion.}
In several Claude Ant artifacts, forward speed is multiplied by heading and uprightness gates, then surrounded by many stability, height, action, joint, and termination penalties.
Although each term is individually plausible, the main progress signal is weakened before a stable gait has emerged.
Early policies may therefore receive little positive progress reward for exploratory movement unless posture and heading are already good.
This is not a criticism of the auxiliary stability terms commonly used in Isaac Lab locomotion tasks. This failure reflects a dependency error between the objective artifact and the reward artifact: auxiliary stability terms are used to gate the primary progress signal, rather than serving as regularizers around it.

\begin{tcolorbox}[title=Observed failure snippet: Ant reward (Claude), colback=artifactred!3!white, colframe=artifactred!60!black, boxrule=0.6pt]
\footnotesize
\[
\begin{aligned}
r_t ={}&
\textcolor{artifactred}{2.0\,v_x^+\,r_{\mathrm{heading}}\,r_{\mathrm{upright}}}
+0.8\,r_{\mathrm{upright}}
+0.5\,r_{\mathrm{heading}}
+0.4\,r_{\mathrm{height}} \\
&\textcolor{artifactred}{-0.30\,|v_y|}
\textcolor{artifactred}{-0.50\,|v_z|}
\textcolor{artifactred}{-0.40\,r_{\mathrm{jump}}}
\textcolor{artifactred}{-4.0\,p_{\mathrm{low\_height}}}
\textcolor{artifactred}{-5.0\,\Ind[\mathrm{early\ termination}]} .
\end{aligned}
\]
The task reward is attenuated by stability gates and surrounded by penalties, so exploratory motion can be weakly rewarded before a stable gait is found.
\end{tcolorbox}

\paragraph{Humanoid locomotion.}
The same pattern is sharper in Claude Humanoid artifacts.
One reward makes forward speed conditional on both uprightness and heading, while adding large penalties for hopping, abnormal height, joint motion, and early termination.
Unlike the successful ARPG Humanoid MDP, where forward velocity remains an additive dense signal, this design may weaken the exploratory signal needed to discover walking.

\begin{tcolorbox}[title=Observed failure snippet: Humanoid reward (Claude), colback=artifactred!3!white, colframe=artifactred!60!black, boxrule=0.6pt]
\footnotesize
\[
\begin{aligned}
r_t ={}&
\textcolor{artifactred}{3.5\,\tanh(v_x^+)\,q_{\mathrm{upright}}\,q_{\mathrm{heading}}}
+1.0\,q_{\mathrm{upright}}
+0.8\,q_{\mathrm{heading}}
+0.5\,q_{\mathrm{height}} \\
&\textcolor{artifactred}{-0.80\,|v_z|}
\textcolor{artifactred}{-0.60\,p_{\mathrm{hop}}}
\textcolor{artifactred}{-8.0\,p_{\mathrm{low\_height}}}
\textcolor{artifactred}{-8.0\,\Ind[\mathrm{early\ termination}]} .
\end{aligned}
\]
Forward velocity is not rewarded as a standalone dense signal; it is multiplied by posture and heading gates and then dominated by large instability penalties.
\end{tcolorbox}

\subsection{Summary}

Overall, these cases show that the failed artifacts are not merely isolated reward-design mistakes.
They reflect broken dependencies among state variables, reward terms, termination events, and simulator-grounded quantities.
This is the failure mode targeted by dependency-aware MDP artifactization: reward and termination logic must remain consistent with the state and action artifacts that make the environment both executable and learnable.

\section{Robustness to Underspecified Task Descriptions}
\label{sec:Underspecified}

In practical usage, users rarely provide task descriptions in a fully specified or formally structured manner.
Instead, natural-language inputs are often ambiguous, incomplete, or expressed at mixed levels of abstraction.
To evaluate the sensitivity of MDP modeling to such input variation, we conduct a case study using an underspecified Humanoid locomotion description, derived from the same setting used in Section~\ref{sec:overall} but stripped of explicit formulation cues.
We use the ARPG configuration available for these runs, instantiated with GPT-5.0, and compare it with monolithic LLM baselines and off-the-shelf agentic systems.

\subsection{Modeling Under Underspecified Task Descriptions}

\begin{wraptable}{r}{0.4\textwidth}
\centering
\caption{MSR under underspecified Humanoid descriptions.}
\label{tab:ablation}
\small
\setlength{\tabcolsep}{2pt}
\begin{tabular}{ccccc}
\toprule
ARPG & Gemini-3 & GPT-5.2 & Claude & Codex \\
\midrule
\textbf{0.8} & 0.3 & 0.0 & 0.0 & 0.7 \\
\bottomrule
\end{tabular}
\end{wraptable}
Table~\ref{tab:ablation} reports MSR over 10 trials with the underspecified Humanoid description.
ARPG maintains a high MSR of $0.8$, suggesting that its dependency-aware modeling process remains stable even when explicit formulation cues are removed.
This result is consistent with the overall Humanoid results in Table~\ref{tab:arpg_comparison}, where ARPG also achieves high modeling success under a more detailed task description.
Thus, the main observation is not that underspecification uniformly degrades performance, but that ARPG remains relatively robust across different levels of input specificity.


The baseline results show a less consistent pattern.
Gemini-3 and Claude remain at the same low MSR levels observed in the detailed-description setting, while GPT-5.2 decreases slightly.
In contrast, Codex improves substantially under the underspecified description.
This suggests that underspecification is not necessarily harder for all LLM-based systems.
In some cases, a shorter and less constrained description may allow a strong general-purpose agentic system to rely on generic locomotion priors, whereas a long, detailed description may introduce additional constraints, distractors, or implementation-specific details that make coherent MDP formulation more difficult.

These results therefore indicate that robustness to natural-language task descriptions should be understood as stability under changes in input specificity, rather than performance under ambiguity alone.
Off-the-shelf systems can sometimes infer plausible MDP artifacts from underspecified inputs, as shown by Codex, but their behavior is not consistent across systems.
ARPG achieves the highest MSR in this setting while also maintaining strong modeling performance in the detailed-description setting, supporting the claim that dependency-aware artifactization provides a more stable formulation process.
Overall, this analysis reinforces the conclusion of Section~\ref{sec:overall}: reliable policy generation requires not only executable simulator code, but also an MDP formulation process that remains stable as task descriptions vary in detail and specificity.

\subsection{Modeling Correction Dynamics}
\label{app:modeling_correction_dynamics}

This section analyzes the internal behavior of the error-correction mechanism in the modeling phase.
Rather than treating the underspecified Humanoid setting only as a performance benchmark, we use it as a stress case that exposes where uncertainty arises during dependency-aware MDP formulation.
Because the input intentionally removes explicit formulation cues, such as precise objectives, constraints, and reward-relevant variables, the resulting correction traces provide insight into which modeling artifacts require the most refinement.

\begin{figure*}[h]
    \begin{center}
    \centerline{\includegraphics[width=0.6\linewidth]{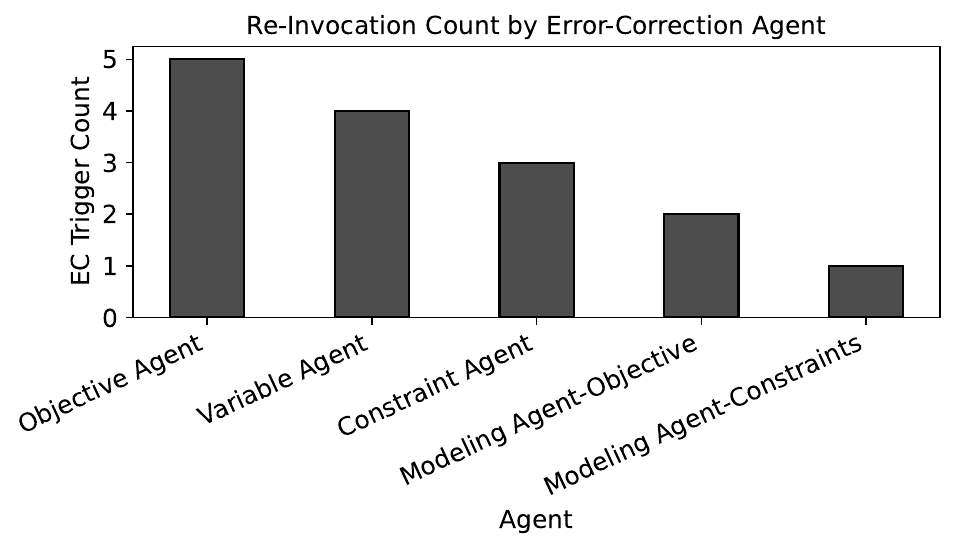}}
    \caption{
    Error-correction agent invocation counts by modeling component with underspecified Humanoid task descriptions.
    \label{fig:app_fig_2}
    }
    \end{center}
\end{figure*}

Figure~\ref{fig:app_fig_2} reports the number of error-correction (EC) agent invocations triggered by each modeling component for the underspecified Humanoid task.
Each bar corresponds to a formulation stage whose artifact was flagged for revision during the sequential modeling process.
A larger count indicates that the corresponding artifact more frequently required additional verification or refinement before being passed to later stages.

The correction traces show that EC invocations are concentrated in the objective and variable agents. This suggests that, under ambiguous task descriptions, the main uncertainty lies in high-level semantic interpretation: determining what constitutes successful behavior and which state variables are needed to represent progress toward that behavior. In contrast, constraint-related artifacts trigger fewer corrections, suggesting that feasibility conditions and physical limits can often be inferred more directly from the robot specification and simulator priors.

Later-stage modeling components require fewer corrections than the earlier semantic interpretation stages. This pattern is consistent with the dependency-aware design of ARPG. Once objectives and variables are externalized and refined, downstream formalization stages operate on more explicit artifacts, reducing ambiguity in the construction of mathematical and SAR-level representations.

Overall, these traces show that modeling difficulty under underspecified input is not uniformly distributed across the MDP formulation pipeline. Uncertainty is concentrated in objective definition and variable selection, while later formalization and constraint encoding are comparatively more stable. This supports the role of stage-wise verification and localized correction: the framework can refine fragile upstream artifacts without regenerating the entire formulation. The observation is also consistent with the theoretical analysis in Section~\ref{subsec:theoretical_justification}, which links localized correction and reduced error propagation to higher modeling success.

{\color{revisionblue}
\paragraph{Pre/post audit of triggered corrections.}
\label{app:ec_audit}
To examine whether EC repairs errors or introduces new ones, we retrospectively audited 50 Humanoid generation trials. EC was triggered in 60 of 350 possible stage events. Each of the 60 triggered EC events used a seven-item checklist, yielding 420 paired pre/post item decisions.

\begin{table}[htbp]
\centering
\color{revisionblue}
\caption{\rev{Pre/post checklist outcomes for triggered Humanoid EC events.}}
\label{tab:ec_audit}
\small
\begin{tabular}{lrr}
\toprule
Checklist outcome & Count & Percentage \\
\midrule
Violation corrected ($\mathrm{X}\to\mathrm{O}$) & 130 & 30.95\% \\
Violation unresolved ($\mathrm{X}\to\mathrm{X}$) & 74 & 17.62\% \\
Satisfied item preserved ($\mathrm{O}\to\mathrm{O}$) & 186 & 44.29\% \\
Violation introduced ($\mathrm{O}\to\mathrm{X}$) & 30 & 7.14\% \\
\bottomrule
\end{tabular}
\end{table}

In Table~\ref{tab:ec_audit}, checklist satisfaction increased from $216/420$ (51.4\%) to $316/420$ (75.2\%). Of the 60 triggered events, 10 resolved all violations and 34 reduced the violation count without resolving all violations, giving a net improvement in $44/60$ events (73.3\%). The other 16 events did not reduce the violation count. EC can therefore leave errors unresolved or introduce new violations. This audit is limited to triggered Humanoid events; untriggered artifacts were not independently labeled, so it does not estimate the false-negative rate of confidence-based routing. Item-level percentages are descriptive because checklist items within an artifact are not independent.
}

\section{Implementation Difficulty Analysis via Debug Statistics}
\label{sec:appendix_difficulty}

This section analyzes how implementation difficulty in ARPG varies across tasks with different structural properties.
After the MDP formulation is fixed, the implementation phase translates the formulation into simulator-specific environment code and invokes debugging when execution or integration errors arise.
We therefore use coding-agent debug invocations as a proxy for the corrective effort required to ground a modeled MDP in an executable Isaac Lab pipeline.

Rather than treating implementation difficulty as a simple function of control dimensionality, we examine where debugging effort is spent across simulator components.
This distinction is important because different tasks stress different parts of the implementation interface: low-dimensional tasks may expose strict API assumptions, locomotion tasks may require careful reward and termination grounding, and manipulation tasks may depend heavily on observation construction.

\begin{figure}[h]
    \begin{center}
    \begin{subfigure}{0.48\linewidth}
        \centering
        \includegraphics[width=\linewidth]{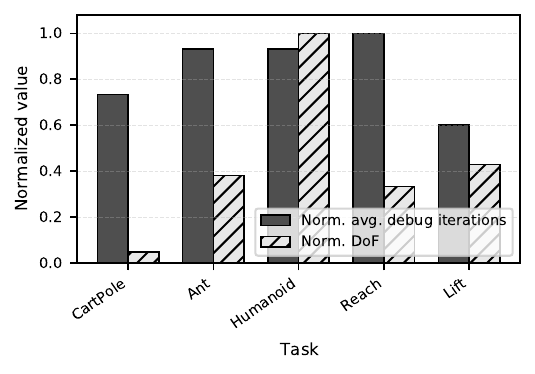}
        \caption{Debug effort vs. DoF.}
        \label{fig:app_fig_1a}
    \end{subfigure}
    \hfill
    \begin{subfigure}{0.48\linewidth}
        \centering
        \includegraphics[width=\linewidth]{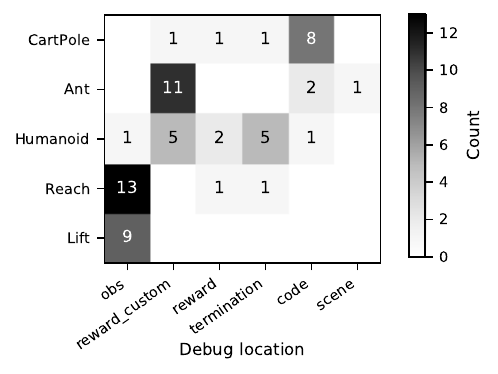}
        \caption{Debug locations by task.}
        \label{fig:app_fig_1b}
    \end{subfigure}
    \caption{
    Implementation difficulty statistics from coding-agent debug invocations.
    \label{fig:app_fig_1}
    }
    \end{center}
\end{figure}

Figure~\ref{fig:app_fig_1a} compares normalized degrees of freedom (DoF) with normalized average debug iterations across the benchmark tasks.
The results show that DoF alone does not explain implementation difficulty.
Humanoid has the largest DoF and also requires substantial debugging effort, but Reach exhibits the highest normalized debug effort despite having a much lower DoF.
CartPole also requires non-negligible debugging effort even though its control dimensionality is minimal, while Lift shows moderate overall effort despite involving object manipulation.
Thus, corrective effort is better explained by simulator-grounding requirements than by action-space dimensionality alone.

Figure~\ref{fig:app_fig_1b} decomposes debug invocations according to the revised implementation component.
Here, \texttt{obs}, \texttt{reward\_custom}, \texttt{termination}, and \texttt{code} denote corrections to observation construction, custom reward implementation, termination logic, and integrated environment code, respectively.
Reach and Lift are dominated by \texttt{obs} corrections, with 13 and 9 invocations, indicating that manipulation tasks are especially sensitive to how object poses, end-effector states, command targets, and relative geometric quantities are exposed to the policy.
Ant is dominated by \texttt{reward\_custom} revisions, with 11 invocations, suggesting that task-specific locomotion rewards require careful grounding in simulator-available quantities beyond the base template.
Humanoid shows a more distributed pattern, with corrections concentrated in both \texttt{reward\_custom} and \texttt{termination}, reflecting the coupling among balance, forward progress, failure detection, and high-dimensional body-state representation.
CartPole, despite its low DoF, concentrates most corrections in \texttt{code}, indicating that even simple control problems can require repeated simulator-interface fixes when the MDP itself is relatively straightforward.

These results suggest that implementation difficulty in ARPG is component-specific rather than uniformly task-specific.
The source of difficulty depends on which part of the generated environment must be grounded most precisely: observation construction for object-centric manipulation, reward implementation for task-specific locomotion, termination logic for unstable high-dimensional systems, and code integration for lightweight environments with strict API assumptions.
Average debug effort and component-level debug location therefore provide complementary signals.
The former estimates total corrective effort, while the latter identifies which implementation artifacts would benefit most from additional validation or more specialized implementation agents.

Overall, this analysis provides a more fine-grained view of ARPG's implementation phase.
When transferring to new simulators or new robot tasks, tasks with rich object-state structure may benefit from stronger observation validation, tasks with custom behavioral objectives may benefit from reward-specific checks, and systems with fragile failure modes may require more explicit termination verification.
This component-level interpretation also clarifies why ARPG separates MDP formulation from simulator implementation: once a valid MDP is fixed, remaining failures are often localized to particular simulator components rather than to the entire policy generation pipeline.

\section{Details of Success Criteria}
\label{app:Details_of_Success}
\subsection{Modeling Success Criteria}
\label{app:modeling_criteria}

This appendix defines the evaluation criteria used to assess whether an MDP
formulated by each model constitutes a valid and meaningful RL
problem. We evaluate MDPs along four components:
\emph{State}, \emph{Action}, \emph{Reward}, and \emph{Overall}.
\rev{An MDP is counted as an MSR success if its trained policy passes PGSR criteria or all applicable component-level criteria are satisfied.}
{\color{revisionblue}
\paragraph{Operational MSR protocol.}
A single researcher with robotics and RL expertise evaluates the realized MDP
with method and model identities removed.
Trials whose trained policies satisfy the PGSR criteria are counted as MSR successes.
For the remaining trials, all applicable criteria in
Tables~\ref{tab:msr_cartpole_protocol}--\ref{tab:msr_lift_protocol}
must pass.
}

\paragraph{\rev{CartPole} Evaluation Criteria.}
An extracted MDP is evaluated using the following implementation-oriented criteria:
\begin{itemize}[leftmargin=*, nosep]
  \item \textbf{State:}
  Includes all four core continuous state variables--cart position, cart velocity,
  pole angle, and pole angular velocity (notation variants allowed)--such that the
  Markov property is naturally satisfied.

  \item \textbf{Action:}
  A single one-dimensional continuous control applied exclusively to the cart
  (e.g., force or effort on \texttt{slider\_to\_cart} or equivalent cart actuation).

  \item \textbf{Reward:}
  Aligned with the objective of keeping the pole upright and explicitly specifies
  at least one task-relevant failure or termination condition
  (e.g., exceeding an angular limit and/or a cart position limit).

  \item \textbf{Overall:}
  State, Action, and Reward jointly define a coherent SAR-formatted MDP, and the
  \rev{final formulation or implementation has clearly identifiable state, action, reward, and termination definitions.}
  \item \textbf{Rubric:} \rev{The exact criteria used to assess the MDP are provided in Table~\ref{tab:msr_cartpole_protocol}.}
\end{itemize}

\begingroup
\color{revisionblue}
\footnotesize
\setlength{\tabcolsep}{3pt}
\renewcommand{\arraystretch}{1.08}
\begin{longtable}{p{0.17\linewidth}p{0.44\linewidth}p{0.33\linewidth}}
\caption{\rev{Operational MSR criteria for CartPole.}}
\label{tab:msr_cartpole_protocol} \\
\toprule
\textbf{Component} & \textbf{Pass criterion} & \textbf{Failure criterion} \\
\midrule
\endfirsthead

\toprule
\textbf{Component} & \textbf{Pass criterion} & \textbf{Failure criterion} \\
\midrule
\endhead

\bottomrule
\endfoot

Structure & Observation/state, action, reward, and failure/termination are identifiable and mutually consistent. & A field is missing or contradictory. \\
Cart position & Continuous cart displacement is observed or explicitly derivable from observations. & Position is missing or only a Boolean is given. \\
Cart velocity & Continuous cart velocity is observed or explicitly derivable from observations. & Velocity is unavailable. \\
Pole angle & Pole angle relative to upright is observed or equivalently represented. & Pole angle is unavailable or only an upright Boolean is given. \\
Pole angular velocity & Continuous pole angular velocity is observed or explicitly derivable. & Angular velocity is unavailable. \\
Action type & Continuous force/effort control has explicit numerical or symbolic bounds. & Control is discrete or its type/bounds are unresolved. \\
Action dimension & One declared schema: 1D cart force. & Cart force is absent or dimensions conflict. \\
Upright objective & The task reward favors smaller pole-angle error, with other inputs fixed. & The upright objective is absent or reversed. \\
Failure boundary & At least one finite pole-angle or cart-position failure boundary is defined. & Only timeout is specified, or the boundary is unresolved. \\
Joint coherence & State, action, and failure logic refer to consistent cart/pole joints. & Joint mappings conflict. \\
\end{longtable}
\endgroup

\paragraph{Ant Evaluation Criteria.}
An extracted MDP is evaluated using the following implementation-oriented criteria:
\begin{itemize}[leftmargin=*, nosep]
  \item \textbf{State:}
  Includes a sufficient set of proprioceptive variables for quadruped locomotion,
  such as base orientation or uprightness cues, base linear and angular velocities,
  and joint positions and velocities for all eight actuated joints.
  In addition, the state (or reward logic) must include information relevant to
  detecting collapse, such as the base height.
  Missing essential proprioceptive components or collapse-related signals is
  considered invalid.

  \item \textbf{Action:}
  An eight-dimensional continuous control corresponding to joint-level effort or
  torque commands applied to the ant's actuated leg joints.
  Actions must be conceptually consistent with joint-effort actuation in Isaac Lab.
  Discrete controls, incorrect dimensionality, or ambiguous actuation semantics
  are considered invalid.

  \item \textbf{Reward:}
  Aligned with the objective of stable forward locomotion along the positive
  X-axis and includes explicit incentives for forward progress as well as penalties
  related to body stability (e.g., tilt or angular velocity) and control efficiency
  (e.g., effort magnitude or action smoothness).
  The reward formulation must also clearly specify termination conditions,
  including collapse due to insufficient base height and a finite
  episode horizon. Rewards that omit or ambiguously define these
  termination conditions are considered invalid.
  \item \textbf{Rubric:} \rev{The exact criteria used to assess the MDP are provided in Table~\ref{tab:msr_ant_protocol}.}
\end{itemize}

\begingroup
\color{revisionblue}
\footnotesize
\setlength{\tabcolsep}{3pt}
\renewcommand{\arraystretch}{1.08}
\begin{longtable}{p{0.17\linewidth}p{0.44\linewidth}p{0.33\linewidth}}
\caption{\rev{Operational MSR criteria for Ant.}}
\label{tab:msr_ant_protocol} \\
\toprule
\textbf{Component} & \textbf{Pass criterion} & \textbf{Failure criterion} \\
\midrule
\endfirsthead

\toprule
\textbf{Component} & \textbf{Pass criterion} & \textbf{Failure criterion} \\
\midrule
\endhead

\bottomrule
\endfoot

Structure & Observation/state, action, reward, collapse, and timeout are identifiable. & A required field is missing. \\
Base state & Orientation/uprightness, height, and 3D linear/angular velocity have explicit raw or derived sources. & Essential base-state information is unavailable. \\
Joint state & Positions and velocities cover all eight actuated joints. & A required joint or its velocity is missing. \\
Action dimension & One continuous 8D action vector is fully specified. & The action is noncontinuous, or its dimension is unresolved or differs from eight. \\
Action semantics & Joint effort/torque or normalized commands map explicitly to the effort actuators. & Commands conflict with the declared actuator interface. \\
Forward signal & Global $+X$ progress or speed is rewarded as a locomotion objective. & Forward progress is absent, direction is unresolved, or progress is only indirectly gated. \\
Stability & Uprightness, tilt, angular stability, or height is shaped in the correct direction. & A stability incentive is absent or reversed. \\
Efficiency & Effort, action magnitude/change, or energy is penalized. & A control-efficiency term is absent or reversed. \\
Collapse & A resolved base-height collapse threshold is at least $0.31$\,m and distinguishes collapse from the declared upright configuration. & The threshold is unresolved, below the benchmark lower bound, or inconsistent with the upright configuration. \\
Horizon & A resolved finite horizon lies in $[16,20]$\,s. & The horizon is absent, unresolved, or outside the benchmark range. \\
Forward frame & Forward motion uses world $+X$ or an explicit body-to-world transform. & Body-frame velocity is used as world-frame forward velocity without a transform. \\
Temporal source & Any previous action used has an observation or declared manager-memory source. & A temporal dependency has no available source. \\
\end{longtable}
\endgroup

\paragraph{Humanoid Evaluation Criteria.}
An extracted MDP for the humanoid locomotion task is evaluated using the following
implementation-oriented criteria:
\begin{itemize}[leftmargin=*, nosep]
  \item \textbf{State:}
  Includes \rev{base position or height and orientation} together with linear and angular velocities,
  as well as joint positions and joint velocities. The state also contains at least
  one stability-relevant signal, such as uprightness, forward heading (+X),
  base height, or foot contact.

  \item \textbf{Action:}
  A continuous per-joint control vector compatible with Isaac Lab's implicit actuator
  model (e.g., joint position targets or normalized joint commands).

  \item \textbf{Reward:}
  Encourages forward progression and stable upright locomotion, and explicitly
  specifies both termination conditions: \rev{a finite episode horizon and a base-height collapse threshold.}

  \item \textbf{Overall:}
  State, Action, and Reward jointly define a coherent SAR-formatted MDP that can be
  used consistently across episodes.
  \item \textbf{Rubric:} \rev{The exact criteria used to assess the MDP are provided in Table~\ref{tab:msr_humanoid_protocol}.}
\end{itemize}

\begingroup
\color{revisionblue}
\footnotesize
\setlength{\tabcolsep}{3pt}
\renewcommand{\arraystretch}{1.08}
\begin{longtable}{p{0.17\linewidth}p{0.44\linewidth}p{0.33\linewidth}}
\caption{\rev{Operational MSR criteria for Humanoid.}}
\label{tab:msr_humanoid_protocol} \\
\toprule
\textbf{Component} & \textbf{Pass criterion} & \textbf{Failure criterion} \\
\midrule
\endfirsthead

\toprule
\textbf{Component} & \textbf{Pass criterion} & \textbf{Failure criterion} \\
\midrule
\endhead

\bottomrule
\endfoot

Structure & Observation/state, action, reward, collapse, and timeout are identifiable and consistent. & A field is missing or the schema is contradictory. \\
Base state & Base position/height, orientation, and linear/angular velocities have explicit raw or derived sources. & Required pose or velocity information is unavailable. \\
Joint state & Positions and velocities cover every controlled joint. & A controlled joint or its velocity is missing, or the state and action joint sets disagree. \\
Stability cue & At least one uprightness, heading, height, or foot-contact cue has a declared raw or derived source. & No stability cue is available. \\
Action schema & One continuous per-joint vector has a uniquely determined dimension. & The action is noncontinuous, its dimension is unresolved, or conflicting schemas remain. \\
Action semantics & Position targets, deltas, or normalized targets map to the implicit actuators. & Raw effort-only commands or other semantics conflict with the actuators. \\
Forward signal & Global $+X$ progress or velocity is rewarded as a locomotion objective. & The reward encourages standing only or forward progress is only indirectly gated. \\
Stability & Uprightness, height, angular stability, or contact provides balance shaping. & Balance shaping is absent or reversed. \\
Collapse & An explicit base-height collapse threshold is at least $0.60$\,m and distinguishes collapse from the declared upright configuration. & The threshold is unresolved, below the benchmark lower bound, or inconsistent with the upright configuration. \\
Horizon & A resolved finite horizon lies in $[16,20]$\,s. & The horizon is absent, unresolved, or outside the benchmark range. \\
Recoverability & Recoverable posture/contact deviations are distinguished from catastrophic termination. & Minor recoverable non-foot contact is unconditionally terminal. \\
Temporal source & Any previous action/action change has an observation or declared manager-memory source. & A temporal dependency has no available source. \\
\end{longtable}
\endgroup

\paragraph{Reach Evaluation Criteria.}
An extracted MDP for the end-effector reaching task is evaluated using the following
implementation-oriented criteria:
\begin{itemize}[leftmargin=*, nosep]
  \item \textbf{State:}
  Includes the robot joint positions and joint velocities, as well as the commanded
  end-effector \rev{position, with orientation and temporal information included as required by the reward formulation.}

  \item \textbf{Action:}
  Joint-space continuous actions compatible with the implicit actuator formulation,
  controlling the robot arm consistently across all degrees of freedom.

  \item \textbf{Reward:}
  Encourages accurate tracking of the commanded end-effector pose, including
  position and (optionally) orientation tracking, and explicitly specifies a
  time-based termination condition.

  \item \textbf{Overall:}
  State, Action, and Reward jointly define a coherent SAR-formatted MDP that can be
  used consistently across episodes.
  \item \textbf{Rubric:} \rev{The exact criteria used to assess the MDP are provided in Table~\ref{tab:msr_reach_protocol}.}
\end{itemize}

\begingroup
\color{revisionblue}
\footnotesize
\setlength{\tabcolsep}{3pt}
\renewcommand{\arraystretch}{1.08}
\begin{longtable}{p{0.17\linewidth}p{0.44\linewidth}p{0.33\linewidth}}
\caption{\rev{Operational MSR criteria for Franka Reach.}}
\label{tab:msr_reach_protocol} \\
\toprule
\textbf{Component} & \textbf{Pass criterion} & \textbf{Failure criterion} \\
\midrule
\endfirsthead

\toprule
\textbf{Component} & \textbf{Pass criterion} & \textbf{Failure criterion} \\
\midrule
\endhead

\bottomrule
\endfoot

Structure & Observation/state, action, reward, and timeout are identifiable. & A required field is missing. \\
Joint state & Positions and velocities cover all seven arm joints. & A required joint or its velocity is missing. \\
Target command & The commanded 3D end-effector position is in, or explicitly derivable from, realized observations. & The commanded position is unavailable from the realized observations. \\
Orientation & If orientation is rewarded, actual and commanded orientations have declared observable sources or equivalents. & One orientation source is unavailable. \\
Action history & If action rate is rewarded, the previous action has an observation or declared manager-memory source. & The previous action is only optional or has no available temporal source. \\
Action dimension & A continuous 7D arm joint-space command is fully specified. & Cartesian/effort-only control conflicts with the interface. \\
Action semantics & Position targets, deltas, or normalized targets map to the implicit actuators. & Action semantics conflict with the actuator interface. \\
Position tracking & The position-tracking task reward increases as position error decreases, including near the target. & The task incentive is reversed or flat near the target. \\
Orientation metric & If quaternion orientation is rewarded, equivalent $q$ and $-q$ have equal reward. & The orientation metric depends on quaternion sign. \\
Timeout & A finite timeout is resolved in seconds or steps. & Timeout is absent or unresolved. \\
Pose frame & Actual and commanded poses share a frame or have an explicit transform. & World and base coordinates are mixed without a transform. \\
\end{longtable}
\endgroup

\paragraph{Franka Lift Evaluation Criteria.}
An extracted MDP for the Franka lift manipulation task is evaluated using the
following implementation-oriented criteria:
\begin{itemize}[leftmargin=*, nosep]
  \item \textbf{State:}
  Includes the robot joint positions and joint velocities\rev{, including the gripper joints}, the object \rev{pose and velocities} relative to the robot, and the commanded target pose. \rev{Temporal information is available as required by the reward and termination logic.}

  \item \textbf{Action:}
  Joint-space actions consistent with the implicit actuator formulation, where
  actions represent joint position targets, joint deltas, or normalized joint
  commands. Arm and gripper dimensions are defined consistently.

  \item \textbf{Reward:}
  Reflects the full manipulation process, including approach, grasp, lift, and
  maintenance, and explicitly includes a drop-related termination condition
  (e.g., object height falling below $-0.05$\,m).

  \item \textbf{Overall:}
  State, Action, and Reward jointly define a coherent SAR-formatted MDP that can
  be used consistently across episodes.
  \item \textbf{Rubric:} \rev{The exact criteria used to assess the MDP are provided in Table~\ref{tab:msr_lift_protocol}.}
\end{itemize}

\begingroup
\color{revisionblue}
\footnotesize
\setlength{\tabcolsep}{3pt}
\renewcommand{\arraystretch}{1.08}
\begin{longtable}{p{0.17\linewidth}p{0.44\linewidth}p{0.33\linewidth}}
\caption{\rev{Operational MSR criteria for Franka Lift.}}
\label{tab:msr_lift_protocol} \\
\toprule
\textbf{Component} & \textbf{Pass criterion} & \textbf{Failure criterion} \\
\midrule
\endfirsthead

\toprule
\textbf{Component} & \textbf{Pass criterion} & \textbf{Failure criterion} \\
\midrule
\endhead

\bottomrule
\endfoot

Structure & Observation/state, action, staged reward, drop, and timeout are identifiable. & A required field is missing. \\
Robot state & Positions/velocities cover seven arm and two gripper joints. & A required arm or gripper position or velocity is missing. \\
Object position & The object 3D position is observed or explicitly derivable from observations. & It is absent or available only as latent simulator state. \\
Object orientation & Object orientation is observed or equivalently represented. & It is optional only, absent, or latent only. \\
Object linear velocity & The object 3D linear velocity is observed or explicitly derivable. & It is absent or latent only. \\
Object angular velocity & The object 3D angular velocity is observed or explicitly derivable. & It is absent or latent only. \\
Relative geometry & Hand/fingertip-to-object geometry is available or computable in a common frame. & Relative geometry cannot be computed consistently. \\
Lift target & The $0.04$\,m lift target relative to initial/table height, or the equivalent commanded target, is explicit. & Lift direction, reference height, or target magnitude is unresolved. \\
Arm action & All seven arm joints receive continuous position targets, deltas, or normalized targets. & Arm dimensions are incomplete or commands conflict with the interface. \\
Gripper action & A two-finger mapping completes one consistent 9D action vector. & Gripper control is absent or 8D/9D schemas conflict. \\
Approach & Reducing approach distance provides a task incentive before grasping. & All task incentives vanish in the pre-grasp region. \\
Grasp acquisition & Proximity, centering, aperture, or contact supplies a continuous grasp-acquisition signal. & Only binary grasp success is rewarded. \\
Lift & Lifting is rewarded through object height or progress. & Only end-effector height is rewarded. \\
Stage bridge & A continuous approach/grasp signal bridges any hard grasp gate on the lift reward. & All task gradients vanish before a hard grasp gate. \\
Maintenance & Grasp/height/hold reward continues at the target height. & Reward occurs only once at threshold crossing. \\
Drop termination & Both timeout and object-drop termination are defined. & Drop has only a penalty without termination, or timeout is absent. \\
Drop frame & The drop threshold has an explicit table/reset/object-frame reference. & The threshold frame or scene anchor is unresolved. \\
Temporal source & Initial height, previous action, or a phase latch has an observation or declared manager-memory source when used. & A temporal dependency has no available source. \\
\end{longtable}
\endgroup

\FloatBarrier
\subsection{Task-Specific Policy Generation Success Criteria}
\label{app:task_thresholds}

For policy generation evaluation, we define task-specific success criteria
in behavioral terms.
Each criterion specifies the observable conditions under which a learned
policy is considered to have successfully completed the intended task.
These criteria are used solely for evaluation and do not affect reward design,
termination conditions, or training dynamics.
\rev{Each trained policy is evaluated automatically on 20 fixed-seed episodes
and passes if at least 16 satisfy all task-specific conditions.
PGSR is computed over all 10 independent generation-and-training trials,
including failures at earlier stages.}

\paragraph{CartPole.}
\rev{An evaluation episode is successful if it completes the 5-s horizon without failure, keeps the cart within 3 m of the track center throughout the episode, and maintains an absolute pole-angle error of at most 0.20 rad throughout the interval from 1 s to 5 s.}

\paragraph{Ant.}
\rev{An evaluation episode is successful if it completes the full 16-s horizon without early termination, never lets the base height fall below 0.31 m, and achieves a mean forward velocity of at least 0.50 m/s. Throughout the interval from 1 s to 16 s, the up-axis projection must remain above 0.93 and the heading projection above 0.80.}

\paragraph{Humanoid.}
\rev{An evaluation episode is successful if it completes the full 16-s horizon without early termination, never lets the base height fall below 0.60 m, and achieves a mean forward velocity of at least 0.50 m/s. Throughout the interval from 1 s to 16 s, the up-axis projection must remain above 0.93 and the heading projection above 0.80.}

\paragraph{Reach.}
\rev{An evaluation episode is successful if, for each of three 4-s targets, the end-effector position error remains at most 0.10 m throughout the final 1-s hold window.}

\paragraph{Lift.}
\rev{An evaluation episode is successful if it completes the full 5-s horizon without failure, never lets the object height fall below -0.05 m, and maintains an object height above 0.04 m and a goal-position error of at most 0.05 m throughout the final second.}

\subsection{\rev{Confidence Intervals for Success Rates}}
\label{app:confidence_intervals}
{\color{revisionblue}
\begin{wraptable}{r}{0.3\textwidth}
\centering
\color{revisionblue}
\vspace{-1.0em}
\caption{\rev{95\% Wilson CIs.}}
\label{tab:wilson_intervals}
\small
\setlength{\tabcolsep}{4pt}
\begin{tabular}{ccc}
\toprule
Successes & Rate & 95\% CI \\
\midrule
0/10 & 0.0 & [0.00, 0.28] \\
1/10 & 0.1 & [0.02, 0.40] \\
2/10 & 0.2 & [0.06, 0.51] \\
3/10 & 0.3 & [0.11, 0.60] \\
4/10 & 0.4 & [0.17, 0.69] \\
5/10 & 0.5 & [0.24, 0.76] \\
6/10 & 0.6 & [0.31, 0.83] \\
7/10 & 0.7 & [0.40, 0.89] \\
8/10 & 0.8 & [0.49, 0.94] \\
9/10 & 0.9 & [0.60, 0.98] \\
10/10 & 1.0 & [0.72, 1.00] \\
\bottomrule
\end{tabular}
\end{wraptable}
All task-wise MSR, EISR, and PGSR values in Table~\ref{tab:arpg_comparison} and the matched ablations are proportions of $n=10$ independent generation/training trials. For a fixed confidence level and sample size, the binomial Wilson interval depends only on the number of successes $k$. Thus, the same reported rate has the same interval across methods, tasks, and metrics. Table~\ref{tab:wilson_intervals} provides all possible intervals without adding columns to the main results tables.

For $\hat p=k/n$ and $z=1.959964$, the two endpoints are
\begin{equation*}
\frac{\hat p+z^2/(2n)\;\pm\;z\sqrt{\hat p(1-\hat p)/n+z^2/(4n^2)}}{1+z^2/n}.
\end{equation*}
The intervals are computed using 10 independent trials for each
method and task, treating each trial outcome as a Bernoulli observation
under the stated evaluation protocol.
\par
}

\FloatBarrier
\section{Additional Generality Demonstrations}
\label{sec:additional_generality}

\subsection{Overview}
To further examine whether ARPG can be instantiated beyond the Isaac Lab tasks used in the main experiments, we provide additional demonstrations on a broader set of sequential decision-making problems.
These demonstrations cover custom robotics tasks, standard RL benchmark tasks, and domain-specific engineering problems.
Across these settings, the dependency-aware modeling pipeline is retained, while only the simulator- or environment-specific implementation interface is adapted to the target domain.
Because the tasks use heterogeneous evidence types, including rollout videos, evaluation curves, quality-of-service (QoS) metrics, and expert comparisons, we report them as generality demonstrations rather than as a unified success-rate benchmark.

\begin{table}[h]
\caption{Summary of additional demonstrations beyond the main Isaac Lab benchmark tasks.}
\label{tab:additional_generality_summary}
\centering
\footnotesize
\setlength{\tabcolsep}{3pt}
\begin{tabular}{
>{\raggedright\arraybackslash}p{0.16\linewidth}
>{\raggedright\arraybackslash}p{0.15\linewidth}
>{\raggedright\arraybackslash}p{0.14\linewidth}
>{\raggedright\arraybackslash}p{0.32\linewidth}
>{\raggedright\arraybackslash}p{0.15\linewidth}
}
\toprule
Task & Task type & Backend & Objective & Reported evidence \\
\midrule
Diagonal CartPole &
Robotics variant &
Isaac Sim &
Balance a cart-pole system with a diagonal initial pole. &
Rollout video \\
Jumping Ant &
Robotics variant &
Isaac Sim &
Perform repeated vertical jumps in place. &
Rollout video \\
In-place rotation Ant &
Robotics variant &
Isaac Sim &
Rotate the Ant in place without translational drift. &
Rollout video \\
Jumping Humanoid &
Robotics variant &
Isaac Sim &
Generate vertical jumping behavior for a humanoid robot. &
Rollout video \\
Pick-and-place &
Robotics variant &
Isaac Sim &
Transfer a cube between two tables. &
Rollout video \\
\midrule
CartPole &
Classic control &
Gymnasium &
Balance the pole with discrete cart forces. &
Evaluation curve \\
MountainCar &
Classic control &
Gymnasium &
Reach the hilltop goal under sparse rewards. &
Evaluation curve \\
Drone delivery &
Grid-world MDP &
Grid env. &
Complete pickups and deliveries before energy depletion. &
Evaluation curve \\
Inventory management &
Stochastic control &
Custom env. &
Choose item-wise orders under demand uncertainty and costs. &
Evaluation curve \\
\midrule
Wireless scheduling &
Engineering MDP &
Engineering env. &
Satisfy QoS while minimizing transmit power. &
Expert comparison (reward \& QoS) \\
Energy management &
Engineering MDP &
Engineering env. &
Satisfy demand using grid energy and battery storage. &
Expert comparison (pattern \& cost) \\
\bottomrule
\end{tabular}
\end{table}

Table~\ref{tab:additional_generality_summary} summarizes the task diversity in these additional demonstrations.
The tasks span additional Isaac Sim robot behaviors, Gymnasium benchmarks, lightweight decision-making environments, and engineering-domain MDPs, thereby testing whether generated MDP artifacts remain useful across different state, action, reward, and constraint structures.
The Isaac Sim robotics tasks use the same implementation-phase structure as the main robotics experiments, including per-configuration implementation agents.
The non-robotic tasks use a simplified implementation phase with a single coding agent.
Thus, the robotics demonstrations provide qualitative evidence of transfer to additional robot behaviors, while the quantitative plots below focus on settings evaluated with the simplified implementation pipeline.

\subsection{Additional Isaac Sim Robotics Tasks}

For the additional Isaac Sim robotics tasks, we provide rollout videos in the supplementary material rather than static figures. These demonstrations verify that ARPG can instantiate new robot behaviors by changing the task description and robot/task configuration without changing the dependency-aware modeling pipeline.

\subsection{Standard RL and Decision-Making Tasks}

We next consider standard RL and lightweight decision-making tasks outside Isaac Sim.
These tasks include classic control benchmarks, grid-based planning, and stochastic inventory control, each with different state, action, and reward structures from the robotics environments.
Unlike the Isaac Sim tasks, they use lightweight Gymnasium or Python-based backends and are implemented with a simplified implementation phase using a single coding agent.
This setting tests whether the dependency-aware modeling pipeline can generate executable MDP formulations when the simulator interface and task dynamics differ substantially from robot control.

\begin{figure}[ht]
\centering
\begin{subfigure}{0.4\linewidth}
    \centering
    \includegraphics[width=\linewidth]{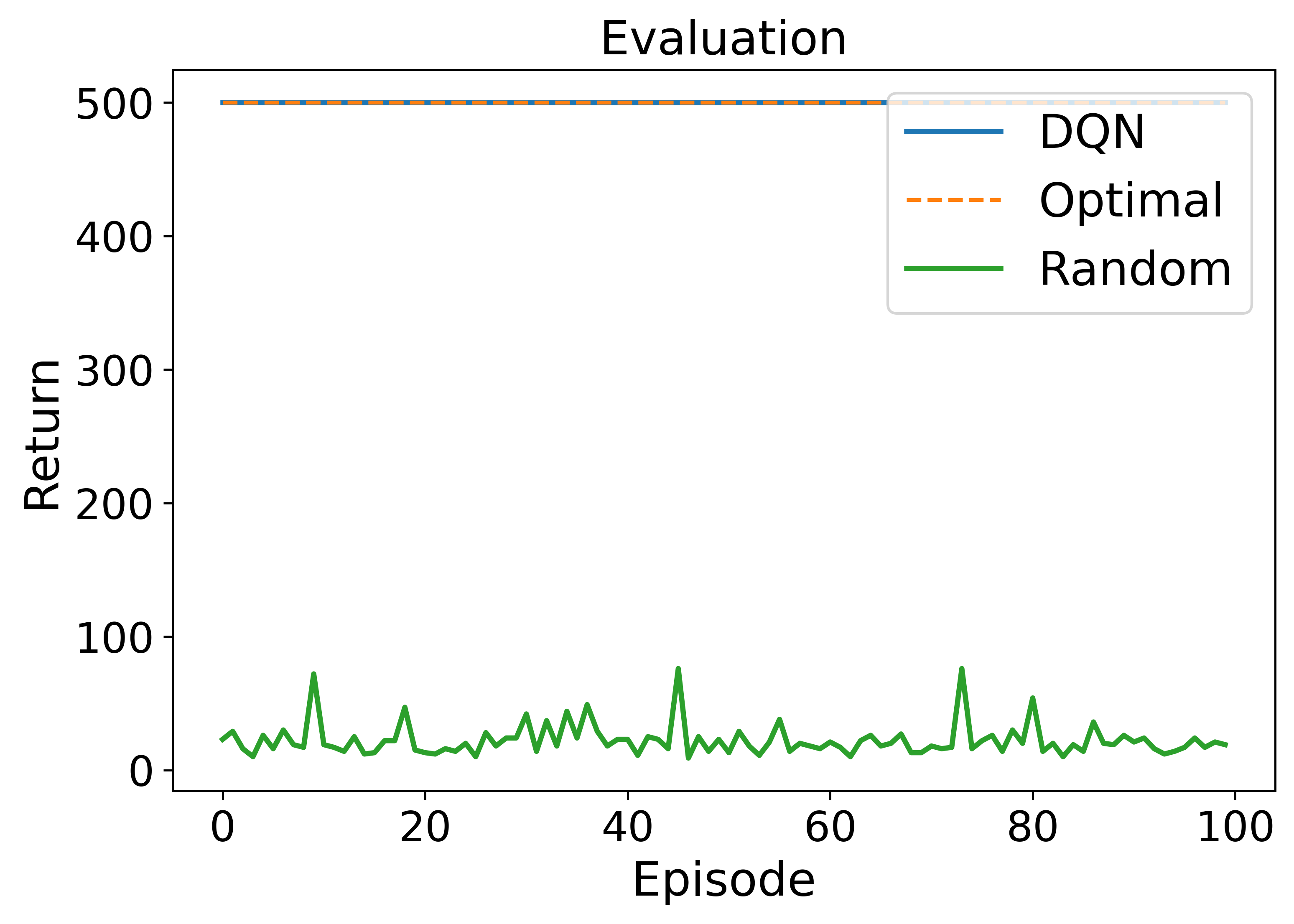}
    \caption{CartPole}
\end{subfigure}
\hfill
\begin{subfigure}{0.4\linewidth}
    \centering
    \includegraphics[width=\linewidth]{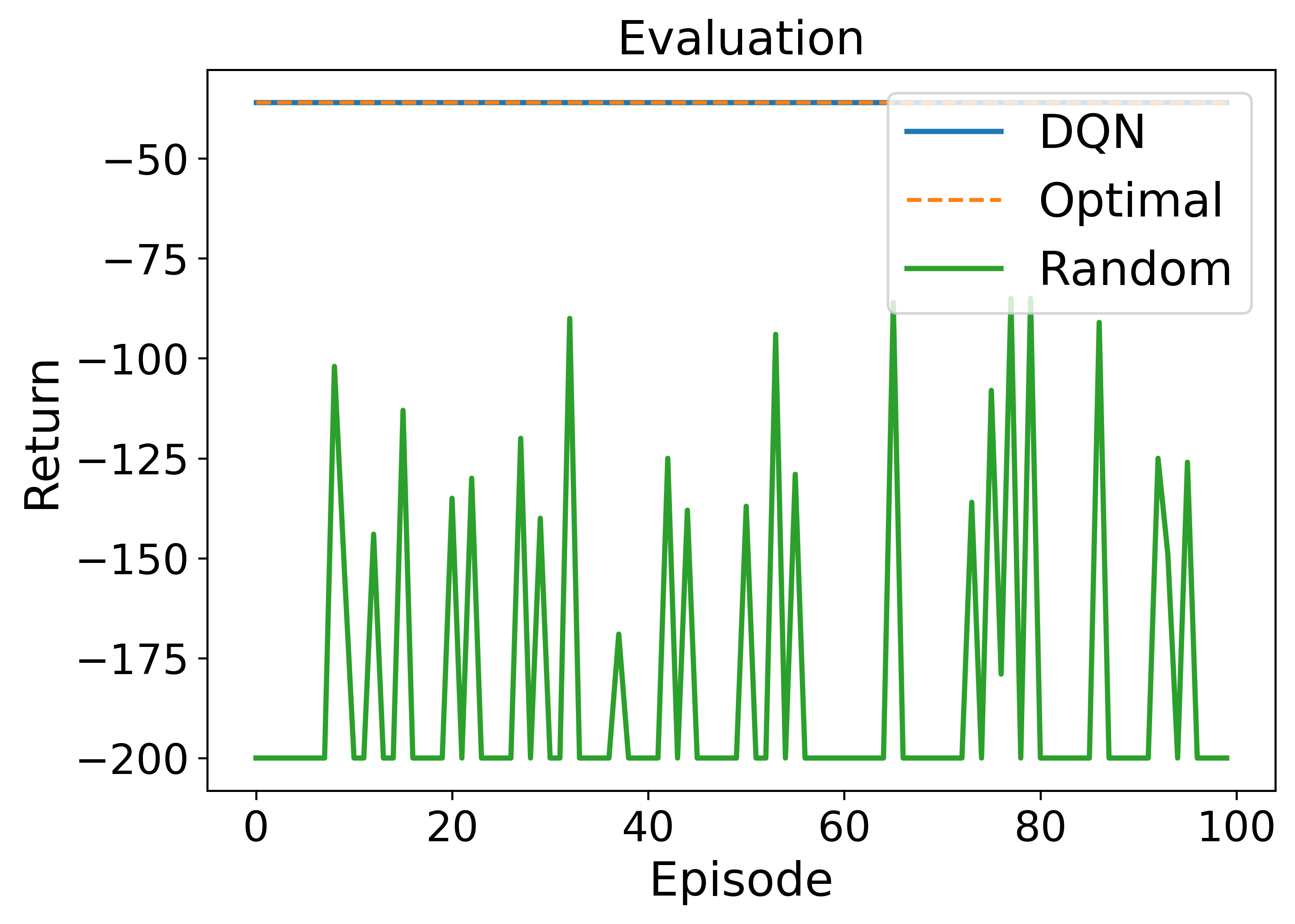}
    \caption{MountainCar}
\end{subfigure}

\vspace{0.5em}
\begin{subfigure}{0.4\linewidth}
    \centering
    \includegraphics[width=\linewidth]{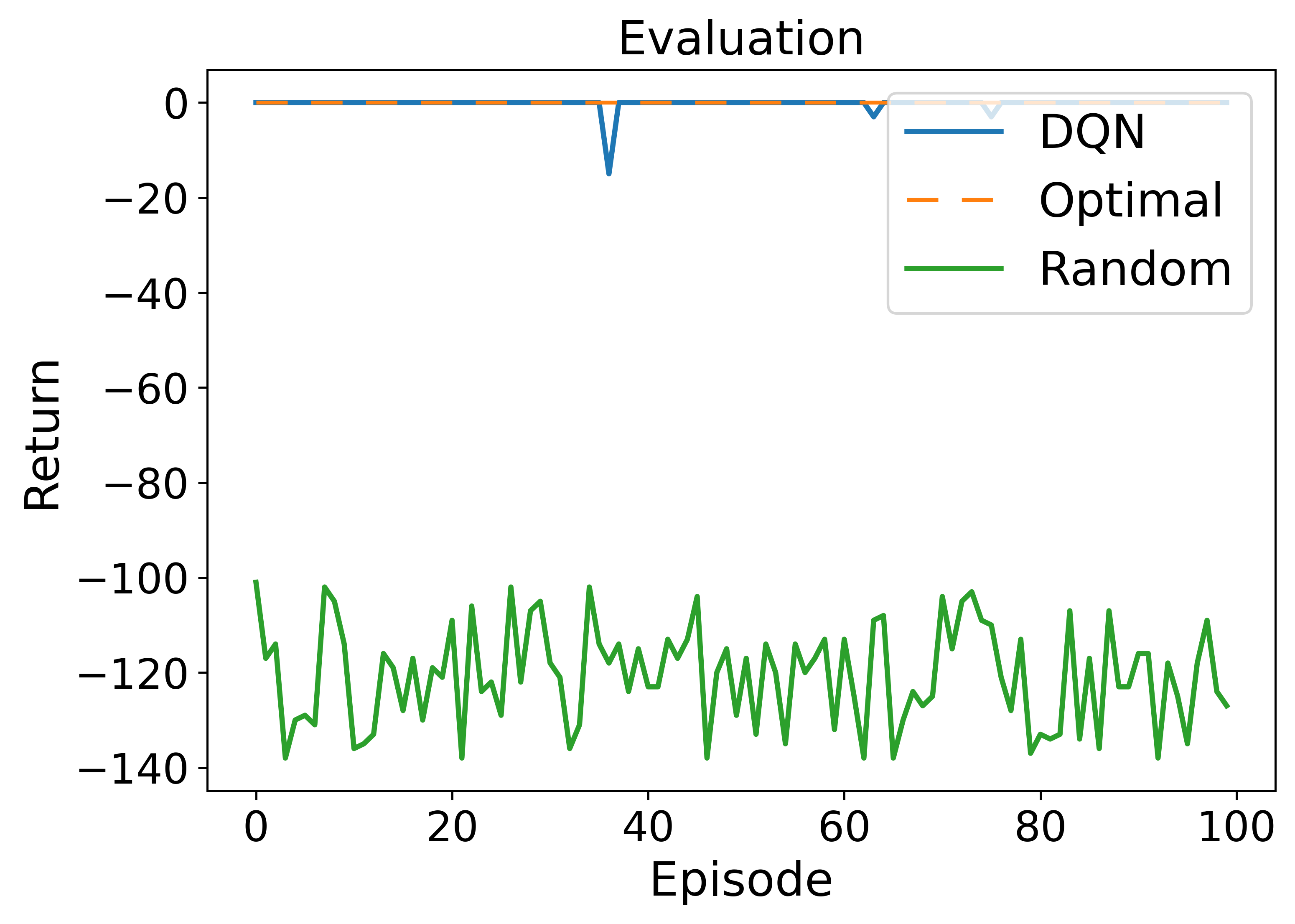}
    \caption{Drone delivery}
\end{subfigure}
\hfill
\begin{subfigure}{0.4\linewidth}
    \centering
    \includegraphics[width=\linewidth]{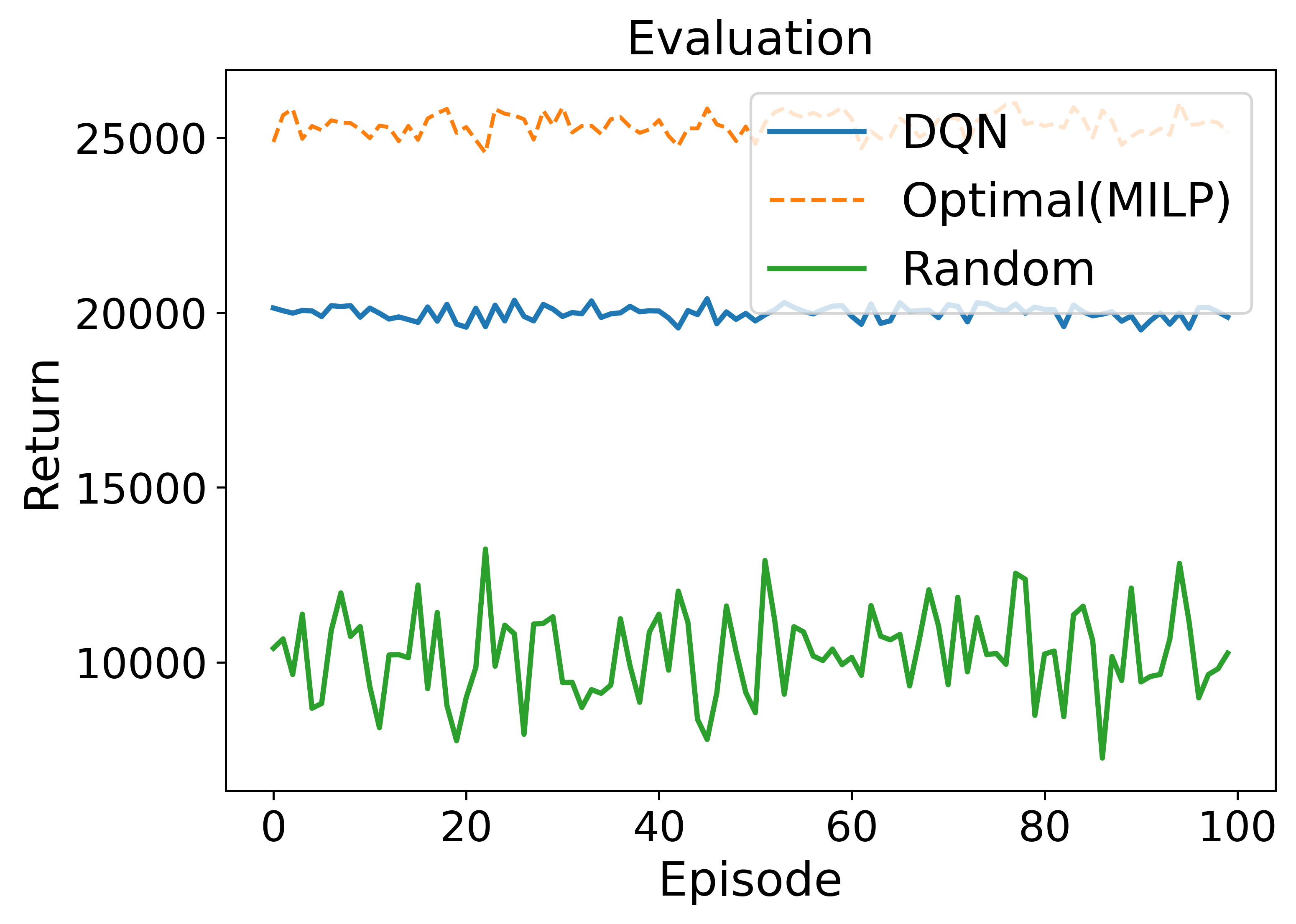}
    \caption{Inventory management}
\end{subfigure}
\caption{Evaluation results for additional standard RL and stochastic decision-making tasks.}
\label{fig:additional_standard_tasks}
\end{figure}

Figure~\ref{fig:additional_standard_tasks} reports evaluation results on four additional standard RL and stochastic decision-making tasks.
These non-Isaac tasks use different simulator interfaces and reward templates from the main robotics experiments.
Nevertheless, ARPG produces executable formulations whose learning curves indicate successful policy optimization, suggesting that the explicit MDP artifact chain is not tied to a single simulator family.

\subsection{Engineering-Domain MDPs}

We consider two engineering-domain MDPs with objectives and constraints that differ substantially from robot locomotion and manipulation.
For both tasks, ARPG uses the same modeling phase agents, while the implementation phase is adapted to lightweight Python environments.

\subsubsection{Wireless Scheduling}

We consider a wireless scheduling problem in which a base station allocates radio resources to multiple users over a sequence of time slots.
At each slot, the controller observes the users' channel conditions and traffic-related state, selects which user to serve, and chooses a transmit-power level.
The objective is to reduce power consumption while maintaining communication quality, expressed through a minimum average-rate requirement.
This task differs from the robotics benchmarks because the dynamics arise from stochastic channel variation and queue/resource-allocation decisions rather than physical robot motion.
It therefore tests whether ARPG can formulate a constrained MDP whose reward and constraints capture a domain-specific trade-off between efficiency and service reliability.

\begin{figure}[h]
\centering
\begin{subfigure}{0.49\linewidth}
    \centering
    \includegraphics[width=\linewidth]{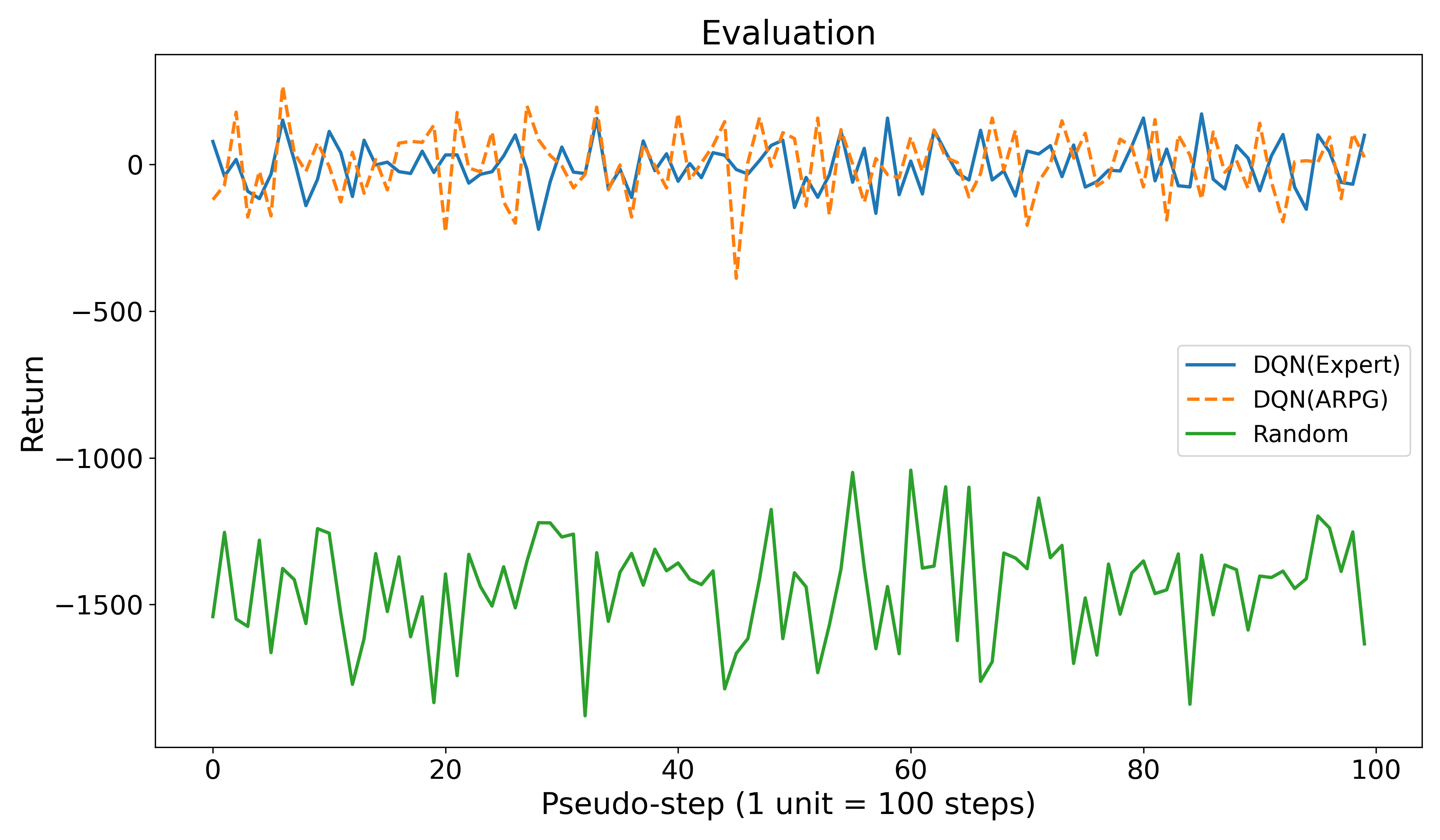}
    \caption{Wireless scheduling reward}
\end{subfigure}
\hfill
\begin{subfigure}{0.49\linewidth}
    \centering
    \includegraphics[width=\linewidth]{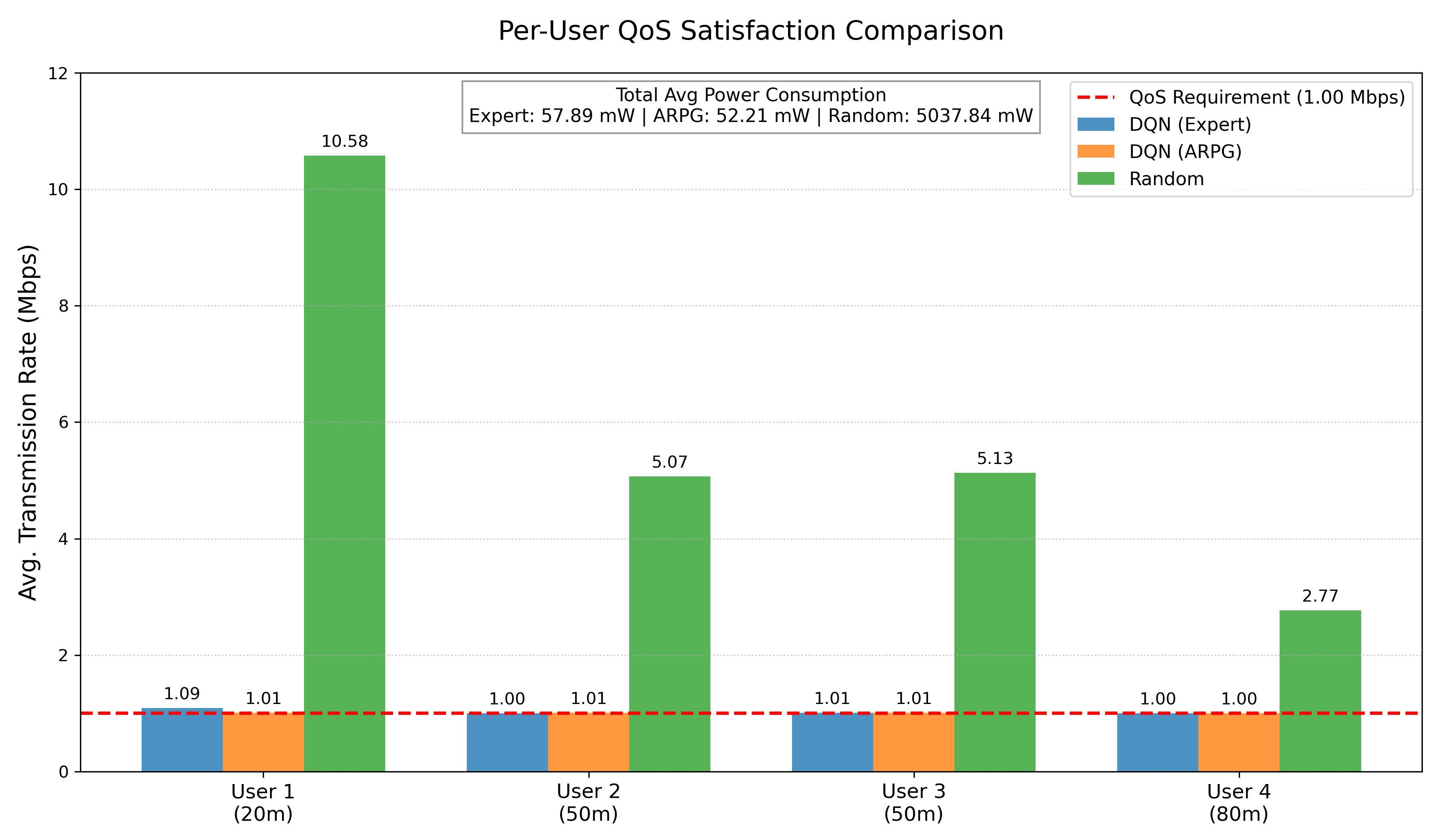}
    \caption{Wireless QoS satisfaction}
\end{subfigure}
\caption{Comparison with the expert-designed DQN policy for the wireless scheduling task.}
\label{fig:additional_wireless_task}
\end{figure}

Figure~\ref{fig:additional_wireless_task} shows the wireless scheduling experiment.
ARPG formulates the task as a constrained MDP in which the agent observes time-varying channel conditions and jointly selects the scheduled user and transmit-power level at each slot.
The ARPG-generated policy reaches reward levels close to those of the expert-designed DQN baseline while satisfying the required QoS constraint throughout evaluation.
The policy also avoids unnecessarily high transmit-power allocation, indicating that the generated reward and constraint artifacts capture the intended trade-off between service reliability and power efficiency.
These results provide evidence that ARPG can generate meaningful MDP formulations for constrained resource-allocation problems beyond physical robot control.

\subsubsection{Energy Management}

We consider an energy management problem in which a controller schedules grid purchase, renewable usage, and battery charging or discharging over a sequence of time steps.
At each step, the controller observes demand, renewable generation, battery state of charge, and time-of-use (TOU) electricity prices.
The objective is to satisfy demand while minimizing total electricity cost, using the battery to shift energy consumption away from high-price periods when possible.
This task differs from the robotics benchmarks because the dynamics are governed by storage constraints, exogenous demand, renewable availability, and time-varying prices rather than physical robot motion.
It therefore tests whether ARPG can formulate a cost-sensitive scheduling MDP with resource-balance constraints and temporally coupled decisions.

\begin{figure}[ht]
\centering
\includegraphics[width=0.65\linewidth]{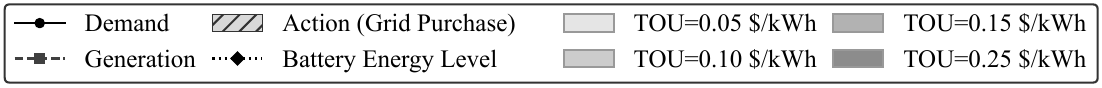}
\begin{subfigure}{0.47\linewidth}
    \centering
    \includegraphics[width=\linewidth]{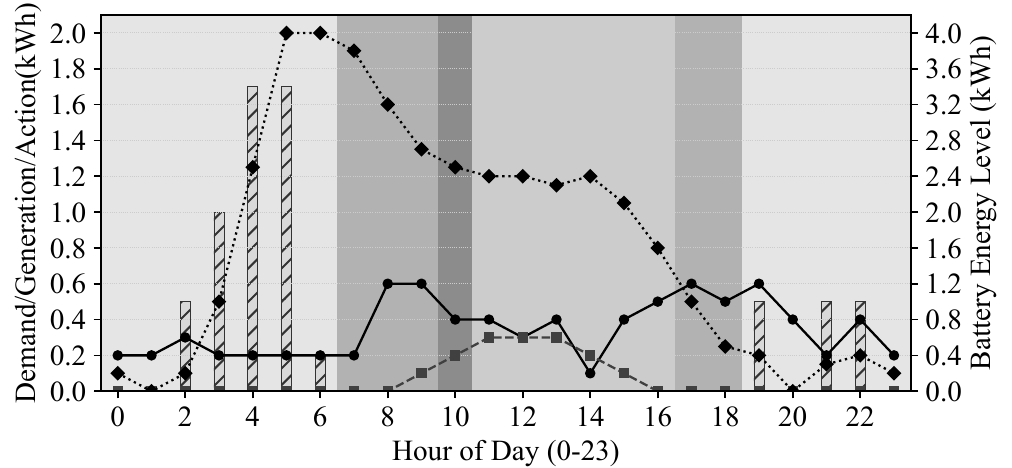}
    \caption{Expert energy-management policy}
\end{subfigure}
\hfill
\begin{subfigure}{0.47\linewidth}
    \centering
    \includegraphics[width=\linewidth]{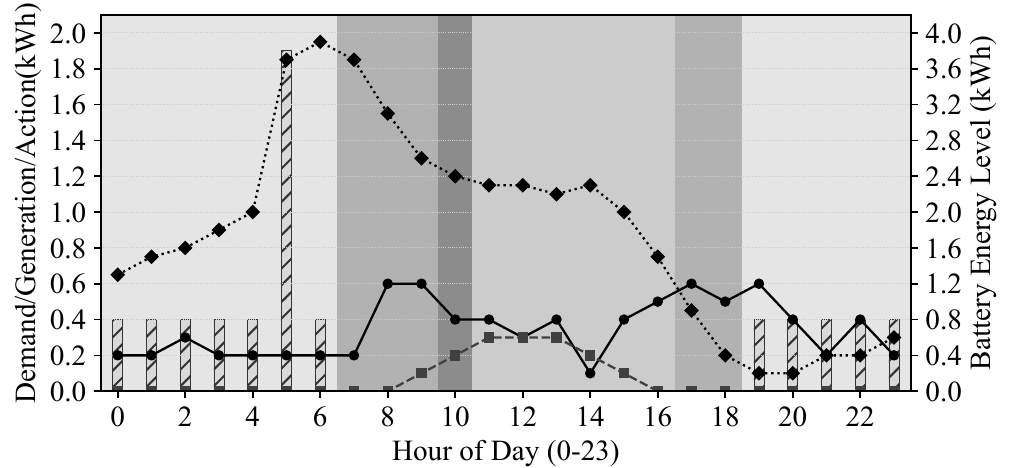}
    \caption{ARPG-generated energy-management policy}
\end{subfigure}
\caption{Comparison of daily patterns with the expert-designed DQN policy for the energy management task.}
\label{fig:additional_energy_task}
\end{figure}

Figure~\ref{fig:additional_energy_task} shows the energy-management experiment.
ARPG formulates the task as a cost-minimization MDP over demand, renewable generation, battery state, and time-of-use prices.
The ARPG-generated policy follows a scheduling pattern similar to the expert-designed policy: it uses renewable energy when available, charges or discharges the battery according to price and demand conditions, and reduces grid purchases during expensive periods while satisfying demand.
This behavior indicates that the generated state, reward, and constraint artifacts capture the intended trade-off among demand satisfaction, storage usage, and electricity cost.

\begin{table}[h]
\caption{Monthly and annual electricity costs for the expert and ARPG-generated
energy management policies.}
\label{tab:additional_energy_costs}
\centering
\footnotesize
\setlength{\tabcolsep}{3pt}
\begin{tabular}{lrrrrrrrrrrrrr}
\toprule
\textbf{Policy} & \textbf{Jan} & \textbf{Feb} & \textbf{Mar} & \textbf{Apr} & \textbf{May} & \textbf{Jun} & \textbf{Jul} & \textbf{Aug} & \textbf{Sep} & \textbf{Oct} & \textbf{Nov} & \textbf{Dec} & \textbf{Total} \\
\midrule
Expert & 13.65 & 11.91 & 9.50 & 9.27 & 7.99 & 10.27 & 19.27 & 20.69 & 11.35 & 9.27 & 10.59 & 13.62 & 147.40 \\
ARPG & 13.63 & 11.86 & 9.52 & 9.15 & 8.00 & 10.31 & 20.05 & 21.11 & 11.39 & 9.27 & 10.64 & 13.75 & 148.72 \\
\bottomrule
\end{tabular}
\end{table}

Table~\ref{tab:additional_energy_costs} reports the monthly and annual electricity costs of the expert-designed and ARPG-generated policies.
The ARPG-generated policy obtains an annual cost of 148.72, close to the expert-designed policy cost of 147.40, corresponding to a 0.90\% relative gap.
This result suggests that ARPG can generate a meaningful MDP formulation for cost-sensitive energy scheduling, a setting whose objectives and constraints differ substantially from standard robot locomotion and manipulation rewards.

\subsection{Summary}
Overall, these demonstrations provide supporting evidence that ARPG's dependency-aware decomposition is not limited to the specific robot embodiments used in the main benchmark.
The main quantitative evaluation remains the Isaac Lab experiments in the main paper, while these additional cases illustrate that explicit objective, state, action, reward, and constraint artifacts can be instantiated across substantially different sequential decision-making domains.

\section{Representative ARPG Logbook Trace of Humanoid Task}
\label{app:representative_logbook_trace}

This section provides a compact representative artifact trace for the Humanoid locomotion task to illustrate the intermediate outputs generated by ARPG.
The trace is shortened for readability and omits repeated low-level simulator fields.
Additional task descriptions, full artifact traces, representative prompt excerpts, per-configuration documents, and implementation details are provided as separate supplementary files.

\begin{tcolorbox}[title=Task summary: Humanoid locomotion]
The humanoid must move forward along the global $+X$ direction while maintaining upright balance.
Each episode lasts 16 seconds and terminates early if the base height falls below $0.6$m.
The policy outputs joint-level commands under action decimation, and observations include proprioceptive and base-state information such as joint states, base velocity, orientation, and height.
The task requires forward progress, heading alignment, upright posture, and stable locomotion under randomized initial conditions.
\end{tcolorbox}

\begin{tcolorbox}[title=Parameters (summary)]
\begin{itemize}[leftmargin=1.2em, nosep]
  \item \textbf{Timing}: physics timestep $\Delta t=1/120$s, action decimation $d=2$, control frequency $\approx 60$Hz.
  \item \textbf{Episode}: horizon $T=16$s; early termination if base height $h_t < 0.6$m.
  \item \textbf{Task frame}: forward direction is global $+X$; upright direction is global $+Z$.
  \item \textbf{Robot}: actuated humanoid joints with joint-specific actuator gains and limits.
  \item \textbf{Environment}: flat ground with randomized initial root pose, root velocity, joint positions, and joint velocities.
\end{itemize}
\end{tcolorbox}

\begin{tcolorbox}[title=Variables (summary)]
\begin{itemize}[leftmargin=1.2em, nosep]
  \item \textbf{Joint state}: joint positions $q_t$ and joint velocities $\dot q_t$ for all actuated degrees of freedom.
  \item \textbf{Action}: joint-level command targets $a_t$ applied through the actuator model.
  \item \textbf{Base state}: base position, orientation, linear velocity, angular velocity, and height $h_t$.
  \item \textbf{Task variables}: forward velocity $v^x_t$, heading error $\theta_t$, uprightness $u_t$, lateral position $y_t$, and lateral velocity $v^y_t$.
  \item \textbf{Contact variables}: optional left/right foot contact indicators and contact forces.
\end{itemize}
\end{tcolorbox}

\begin{tcolorbox}[title=Objective]
Learn a policy $\pi$ that maximizes forward locomotion while maintaining upright posture, heading alignment, and lateral stability:
\[
\max_{\pi}\;
\E_{\pi}
\left[
\sum_{t=0}^{T_c-1}
\left(
r^{\mathrm{fwd}}_t
-
r^{\mathrm{head}}_t
-
r^{\mathrm{upr}}_t
-
r^{\mathrm{lat}}_t
-
r^{\mathrm{fall}}_t
\right)
\right].
\]
Here, $r^{\mathrm{fwd}}_t$ rewards forward velocity, while the penalty terms discourage heading error, non-upright posture, lateral drift, and falling.
\end{tcolorbox}

\begin{tcolorbox}[title=Constraints (summary)]
\begin{itemize}[leftmargin=1.2em, nosep]
  \item \textbf{Timing}: actions are held for $d=2$ physics steps; control horizon is derived from the 16s episode duration.
  \item \textbf{Termination}: terminate if $t \ge T$ or $h_t < 0.6$m.
  \item \textbf{Orientation}: base orientation is represented as a unit quaternion; heading and uprightness are derived from the base frame.
  \item \textbf{Actuation}: normalized joint commands are clipped and mapped to joint-specific actuator targets.
  \item \textbf{Contacts}: foot contact indicators and forces describe ground interaction when available.
\end{itemize}
\end{tcolorbox}

\begin{tcolorbox}[title=State]
The SAR agent constructs a proprioceptive and base-centric state:
\[
s_t =
\big[
q_t,\;
\dot q_t,\;
v_t,\;
\omega_t,\;
\theta_t,\;
u_t,\;
v^x_t,\;
y_t,\;
v^y_t,\;
h_t,\;
c^L_t,\;
c^R_t
\big].
\]
This state contains joint motion, base motion, heading, uprightness, forward progress, lateral deviation, height, and optional foot-contact information.
\end{tcolorbox}

\begin{tcolorbox}[title=Action]
The action is a continuous joint-level command vector:
\[
a_t \in [-1,1]^{n_{\mathrm{dof}}}.
\]
Each action dimension is clipped to $[-1,1]$, scaled to joint-specific actuator ranges, and held constant over the action-decimation interval.
\end{tcolorbox}

\begin{tcolorbox}[title=Reward]
The reward combines forward progress with stability penalties:
\[
r_t =
r^{\mathrm{fwd}}_t
-
\lambda_{\mathrm{head}} r^{\mathrm{head}}_t
-
\lambda_{\mathrm{upr}} r^{\mathrm{upr}}_t
-
\lambda_{\mathrm{lat}} r^{\mathrm{lat}}_t
-
\lambda_{\mathrm{fall}} r^{\mathrm{fall}}_t.
\]
A representative instantiation uses
\[
r^{\mathrm{fwd}}_t = v^x_t,\quad
r^{\mathrm{head}}_t = |\theta_t|,\quad
r^{\mathrm{upr}}_t = 1-u_t,\quad
r^{\mathrm{lat}}_t = |y_t| + |v^y_t|,\quad
r^{\mathrm{fall}}_t = \mathbb{I}\{h_t < 0.6\}.
\]
\end{tcolorbox}

\end{document}